# On the Probabilistic Continuous Complexity Conjecture

Mark A. Kon[1], Boston University

**Abstract:** In this paper we prove the probabilistic continuous complexity conjecture. In continuous complexity theory, this states that the complexity of solving a continuous problem with probability $1 - \delta$ converges (as $\delta$ converges to 0), the complexity of solving the same problem in its worst case. We prove the conjecture holds if and only if space of problem elements is uniformly convex. The non-uniformly convex case has a striking counterexample in the problem of identifying a Brownian path in Wiener space, where it is shown that probabilistic complexity converges to only *half* of the worst case complexity in the $\delta \to 0$ limit.

## 1. Introduction

In this paper we consider a situation in which probabilistic continuous complexity theory can be compared to worst case complexity, where problem elements are restricted to bounded sets. In [TWW] it is conjectured (section 8.5) that probabilisitc complexity of a continuous probelm converges to its worst case complexity as the allowed probability of error $\delta$ approaches 0. Here we prove this conjecture, showing that convergence of probabilistic complexity to worst case complexity generically holds under weak hypotheses. However, there are some basic and surprising situations in which this does not occur, even at the level of information complexity (which provides lower bounds for full complexity).

Specifically, in the approximation of Brownian motion with partial data (under Wiener measure), it turns out that the $\delta \to 0$ limit of probabilistic complexity does *not* always approach worst case complexity (see [Ko] for the original analysis and proof of this fact). In fact, if we ignore a set of Brownian paths of arbitrarily small probability, the maximum error of an approximation algorithm can be cut in half. This can occur whether or not we restrict ourselves to standard information operations (those where information consists of pointwise function values). This counterintuitive phenomenon can only occur in normed linear spaces which are not uniformly convex, such as Wiener space.

We will mention some background and motivation for this problem. Probabilistic complexity is a standard approach in the context of classical discrete complexity theory, where complexities of such problems as primality testing and theorem checking can be significantly

[1]Research partially supported by Air Force Office of Scientific Research and the National Science Foundation

reduced if algorithms are allowed to fail with arbitrarily small probabilities. In continuous complexity, probabilistic approaches have been taken in the context of full information (for example, in the context of finding zeroes of polynomials, see [Sm]), and partial information (e.g., [W], [TWW]).

Let us consider probabilistic complexity in the more specific context of the recovery of functions. Suppose $F$ is a Banach space of functions on a measure space $X$, and $B_f \subset F$ is convex and balanced. Given $f \in F$, we wish to identify $f$ from partial information of the form $Nf = (f(x_1), f(x_2), \ldots, f(x_n))$. More generally, we are interested in information of the form $NF = (L_1(f), L_2(f), \ldots, L_n(f))$, where $L_i$ are linear functionals. From the strict viewpoint of recovering functions, the present paper will study lower bounds (based on purely informational limits) on the errors of algorithms $\phi$ for recovering $f$ from $Nf$, comparing lower bounds in the probabilistic case with lower bounds in the worst case. These lower bounds determined by radii of information. We will then augment these results to ones which analyze full complexities of algorithms for recovering functions (Theorem 12).

We will give a general description of results here, leaving formal definitions to Section 2. Let $S$ be a linear map between two spaces $F$ and $G$. Let $N : F \to \mathbb{R}^n$ (the information operator) be linear. Let $\mu$ be a probability measure with support in a bounded, convex, balanced set $F_0 \subset F$. We study probabilistically the complexity of approximating $S$ with a composition $\phi \circ N$, with $N$ of finite rank. The parameter $\delta$ is a (small) probability with which we are allowed to break a given tolerance $\epsilon$ in the approximation. We investigate when, as $\delta \to 0$, problem parameters (e.g., radius of information, complexity) approach those of the worst case problem, i.e., that requiring $\epsilon$ accuracy in all cases.

This paper partially extends a result of Heinrich [H] in which $F$ consists of functions in a periodic Sobolev space. Heinrich explicitly estimated the $\delta-$probabilistic cardinalities of information for problem elements in a ball of radius $q$, in the limits of large $q$ and small $\delta$. He showed that for $S$ the Sobolev imbedding operator, the $\delta \to 0$ limit of probabilistic complexity for estimating $Sf$ is the same as in the worst case. Other estimates relating worst case and probabilistic complexity have appeared in [TWW] (§8.5).

The results of this paper involve several natural questions related to the above. Initially we study whether the $\delta = 0$ probabilistic model has the same radii of information and complexity as the worst case model. This asks essentially whether sets of measure 0 can make a difference in analytic complexity. That is, do "impossible" sets of functions (sets of probability 0) make a difference in the worst case complexity, in that worst case complexity can be decreased by removal of such functions from consideration? The short answer in general linear settings is "no". More precisely, we show that in linear probabilistic settings (and uniformly convex $F$), the $\delta = 0$ probabilistic setting is identical, with regard to $\epsilon$-cardinality, to the worst case setting. It does not make things better to remove sets of measure 0 for worst case error.

However, remark 2 after the proof of the theorem in Section 8 shows that in some (rather unrealistic) models of dependence of cost on information $N$ and algorithms $\phi$, it is possible to have sets of measure 0 which make a difference in full complexity (as opposed to just information complexity). Theorem 12 below, however, also gives conditions under which we may ignore sets of measure 0 as well in the full complexity setting; these conditions hold when more restricted and natural models of complexity are used.

All results here cover a general class of measures on $F$ (nonvanishing measures) which include orthogonally invariant and Gaussian measures. We assume throughout (without loss) that the bounded, convex, balanced subset $F_0 \subseteq F$ (in which the unknown $f$ is assumed to be) is in fact the unit ball of $F$.

For the discussion of the case $\delta = 0$, we have:

**Theorem 1:** *Let $F$ be a separable Banach space with a nonvanishing probability measure $\mu$ on its unit ball $F_0$. Let $G$ be a uniformly convex Banach space. Let $S$ be bounded and linear from $F$ to $G$. Then the problem of approximating $S$ in the $\delta = 0$ (``almost worst case'') probabilistic setting is equivalent in terms of radius and cardinality of information to the worst case setting.*

This theorem is a corollary to the proof of Theorems 2 and 3 (see below).

Let Rad $A$ denote the radius of a set $A$ in a metric space. We also mention a more basic fact than the above theorem:

**Corollary:** *Let $A$ be a convex set in a uniformly convex Banach space $F$ and $\mu$ be a nonvanishing measure on $F$. Then for any set $E$ of measure $0$, $\mathrm{Rad}\,(A \sim E) = \mathrm{Rad}\,(A)$.*

We will also show that for a fixed information operator $N$, the probabilistic radius of information converges to the worst case radius of information as $\delta \to 0$. Letting $R_\delta^{\mathrm{prob}}(N)$ be the probabilistic radius of information of $N$ (again with allowance for excluding a set of probability $\delta$ consisting of the most difficult problem elements), and $R^{\mathrm{wor}}(N)$ be the worst case radius, we have:

**Theorem 2:** *Under the same hypotheses as in Theorem 1, lower error bounds in the probabilistic setting approximate those in the worst case setting for a fixed adaptive information operator $N$ as $\delta \to 0$:*

$$R_\delta^{\mathrm{prob}}(N) \xrightarrow[\delta \to 0]{} R^{\mathrm{wor}}(N).$$

The proof of this theorem follows easily from the proof of Theorem 3 below.

We define the $n^{th}$ minimal radius of information $R_n^{\mathrm{wor}}$ to be the infimum of $R^{\mathrm{wor}}(N)$ over all (adaptive) information operators $N : F \to \mathbb{R}^d$, with the analogous definition for the $n^{th}$ minimal probabilistic radius $R_{\delta,n}^{\mathrm{prob}}$ as an infimum over radii $R_n^{\mathrm{prob}}(N)$. By $\epsilon$-cardinality we mean the smallest cardinality $n$ for an information operator $N$ for which the radius of information is $\epsilon$ or less. Thus the $\epsilon$-cardinality (as a function of $\epsilon$) is essentially an inverse to the $n^{th}$ minimal radius (as a function of $n$).

We will prove convergence of probabilistic $\epsilon$-cardinality to worst case cardinality, or equivalently, of the minimal probabilistic radius to the minimal worst case radius. The proof is technical, relying on the fact that in a uniformly convex Banach space, a set's radius is essentially supported on finite dimensional subsets.

**Theorem 3:** *For $F, G$, and $\mu$ as in Theorem 1, let $S : F \to G$ be linear and bounded. Then the probabilistic minimal radius of information of cardinality $n$ satisfies $R_\delta^{\mathrm{prob}}(n) \xrightarrow[\delta \to 0]{} R^{\mathrm{wor}}(n)$, where the latter the minimal worst case radius of information of cardinality $n$.*

*Thus the probabilistic $\epsilon$-cardinality of this problem converges to the worst case $\epsilon$-cardinality as $\delta \to 0$, at any value of $\epsilon$ where worst case $\epsilon$-cardinality is continuous in $\epsilon$.*

**Remark:** We show here that we cannot remove the assumption that $G$ is uniformly convex. Consider the case where $F$ is Wiener space, consisting of all continuous functions $f(t)$ on $[0, 1]$ with $f(0) = 0$. We place the Gaussian Wiener measure on $F$, obtaining a canonical probability measure on all Brownian paths. Let $\mu$ be standard Wiener measure, conditioned to the unit ball $F_0$, so that for $A \subset F_0$,

$$\mu(A) \equiv \eta(A)/\eta(F_0),$$

where $\eta$ is the standard Wiener measure.

We use the usual $C^0$ norm on $F$,

$$\|f(t)\|_{L^\infty} = \sup_{t \in [0,1]} |f(t)|. \tag{1.1}$$

Let the space $G$ (in which we will measure error of approximation) also be Wiener space, but with a different norm. For $f \in G$, let

$$\|f\| = \|f\|_{L^\infty} + |f(1/2)|$$

Let $S : F \to G$ be the identity $S(f) = f$ (we view $F$ and $G$ as the same space, using $G$ because it has the error metric (1.1) ( which interests us). The metric in $G$ differentiates among paths $f(t)$ strongly by their values at $1/2$, so that extra weight is given, e.g., to the random walker's position at noon.

We show first that for the (standard) information operator $Nf = f(1/2)$, the radius of information $R_\delta^{\text{prob}}(N) = 1$ for all $\delta > 0$, while $R_\delta^{\text{wor}}(N) = 2$, showing that the probabilistic radius need not approach the worst case radius for a fixed information operator. To do this we first evaluate the worst case radius. Thus

$$R^{\text{wor}}(N) = \inf_{\phi : \mathbb{R} \to G} \sup_{f \in F_0} \|Sf - \phi(Nf)\| \tag{1.2}$$

$$= \inf_{\phi : \mathbb{R} \to G} \sup_{f \in F_0} \|f - \phi(y)\|$$

$$= \inf_{\phi : \mathbb{R} \to G} \sup_{f \in F_0} \|f - \phi(y)\|_{L^\infty} + |f(1/2)) - \phi(y)(1/2)|,$$

where above $y \equiv (1/2)$. Above, $\phi(y)$ is a function (our best estimate of $f$) which we evaluate at $t = 1/2$.

We now show the above infimum is 2. Consider the value of the expression on the right side of (1.2) for functions $f$ with $f(1/2) = 1$. Then whatever the function $(\phi(y))(\cdot) = (\phi(1))(\cdot)$, the value of the expression whose supremum is on the right of (1.10)(1.2) can be made arbitrarily close to two, using such $f \in F_0$. Indeed, the first term $\|f - \phi(y)\|_{L^\infty}$ can be made arbitrarily close to $1 + |\phi(y)(1/2)|$ given the right $f$ (still with $f(1/2) = 1$). On the other hand, the second term in (1.2) for the same $f$ equals $|1 - \phi(y)(1/2)| \geq 1 - |\phi(y)(1/2)|$, yielding that (1.2) is at least 2, so that $R^{\text{wor}}(N) \geq 2$. It is not hard to see that in fact

$$R^{\text{wor}}(N) = 2. \tag{1.2}$$

We will show below that

$$R^{\text{prob}}_\delta(N) = 1$$

for *any* choice of $\delta > 0$, so that here deleting a set of arbitrarily small measure can reduce the radius of information by a factor of one half.

We now extend the above result on the worst case radius, showing that for *all* information operators $N$ of cardinality 1, the worst case radius $R^{\text{wor}}(N)$ is 2. Thus let $N$ be an arbitrary bounded linear functional on $F$. Then again we have

$$R^{\text{wor}}(N) = \inf_{\phi: \mathbb{R} \to G} \sup_{f \in F_0} \|Sf - \phi(Nf)\|$$

$$= \inf_{\phi: \mathbb{R} \to G} \sup_{f \in F_0} \|f - \phi N f\|_\infty + |f(1/2)) - (\phi N f)(1/2)|,$$

where above $y \equiv f(1/2)$. Note that $\phi N f$ is a continuous function (best estimate of $f$), which we again evaluate at $1/2$ in the second term above. We now show the above infimum is 2. For note $C_0^0[0,1]$ is a closed subspace of $C^0[0,1]$, so that by the Hahn Banach theorem the dual $C_0^0[0,1]^*$ is contained in the Borel measures on $[0,1]$. Thus for any $N$ there is a Borel measure $\mu$ such that

$$Nf = \int f \, d\mu.$$

For fixed $\mu$, we will now show that for any small $\eta > 0$, there is a function $f_\eta$ such that $f_\eta(1/2) = 1$, $f(t) = -1$ for some $t$ with $|t - 1/2| < \eta$, and $\int f_\eta \, d\mu$ is independent of $\eta$. Indeed, if $\mu$ has no point mass at $t = 1/2$, then

$$\mu[1/2 - \eta, 1/2 + \eta] \underset{\eta \downarrow 0}{\downarrow} 0$$

and so $\int_{1/2-\eta}^{1/2+\eta} f \, d\mu$ becomes arbitrarily small, uniformly in $f$, as $\eta \to 0$. Thus no matter what the behavior of $f$ in $[1/2 - \eta, 1/2 + \eta]$ $f$ can be adjusted outside this interval so as to compensate in such a way that $\int f_\eta \, d\mu$ is independent of $\eta$. On the other hand, if $\mu$ has a point mass of weight $c < 1$ at $t = 1/$, then

$$\int f_\eta \, d\mu = cf(1/2) + \int f_\eta \, d\mu' = c + \int f_\eta \, d\mu',$$

where $\mu'$ is absolutely continuous at $t = 1/2$, so that the same argument works here, this time applied to the second term on the right. This shows again that $Nf_\eta$ is independent of $\eta$ in this case. Thus in both of these cases and for any $\phi$,

$$\sup_{f \in F_0} \|f - \phi N f\|_{L^\infty} + |f(1/2) - (\phi N f)(1/2)|$$

$$\geq \|f_\eta - \phi N f_\eta\|_{L^\infty} + |1 - (\phi N f_\eta)(1/2)|. \tag{1.3}$$

Note $\phi N f_\eta$ is independent of $\eta$, so if we choose $\eta$ to be sufficiently small, $\|f_\eta - \phi N f_\eta\|_{L^\infty}$ can be made arbitrarily close to $|(-1) - \phi N f_\eta(1/2)|$ or a larger number, since

$$f_\eta(t_1) = -1$$

for some $t_1$ with $|t_1 - 1/2| \leq \eta$, while $\phi N f_\eta$ is a fixed continuous function, so for $\eta$ small, $\phi N f_\eta(t_1)$ is arbitrarily close to $\phi N f_\eta(1/2)$. Therefore, the expression in the supremum on the left of (1.3) can be made arbitrarily close to $|1 + (\phi N f_\eta)(1/2)| + |1 - (\phi N f_\eta)(1/2)| = 2$, so that by (1.3), for any linear functional $N$ not completely supported at $t = 0$,

$$R^{\text{wor}}(N) = 2. \tag{1.4}$$

If $\mu$ is a unit point mass at $t = 1/2$, then (1.4) has already been proved in (1.2), so that (1.4) holds in all cases.

We now show that the probabilistic radius $R_\delta^{\text{prob}}(N)$ is half of the worst case radius for every $\delta > 0$. Thus define the family of sets

$$F(m) = \{f \in F_0 \mid |f(t) - f(1/2)| < 1 \text{ for } |t - 1/2| \leq 1/m\}.$$

This is very weak Lipschitz-type condition on $f \in F$. Since $F_0$ is a family of continuous functions,

$$F(m) \underset{m \to \infty}{\uparrow} F_0.$$

so for *any* measure $\nu$ on $F_0$ (Gaussian or otherwise),

$$\nu(F(m)) \underset{m \to \infty}{\uparrow} \nu(F_0).$$

Now consider radii of information for $Nf = f(1/2)$ with respect to this family. For any $A \subset F$, define for an $N : F \to \mathbb{R}^d$

$$R(N, SA) \equiv \sup_{y \in \mathbb{R}^d} \text{Rad}(S(N^{-1}(y) \cap A)).$$

Then for any $m$:

$$R(N, SF(m)) = \sup_{y \in \mathbb{R}} \text{Rad}(S(N^{-1}(y) \cap F(m)))$$

$$= \sup_{y \in [-1,1]} \text{Rad}\{f \in S(F(m)) : f(1/2) = y\}$$

$$= \sup_{y \in [-1,1]} \text{Rad}\{f \in SF_0 : f(1/2) = y \text{ and } |f(t) - y| < 1 \text{ for } |t - 1/2| \leq 1/m\}.$$

But for any $y \in [-1, 1]$, one can check that the set whose radius is computed above has a center $c_{ym}(t) \in F_0$ such that

$$c_{ym}(1/2) = y, \qquad c_{ym}(t) = 0 \text{ for } |t - 1/2| \geq 1/m,$$

and $c_{ym}$ decreases (increases) monotonically to 0 as $|t - 1/2|$ increases. Thus for a $c_{ym}$ with the above properties we have for $f \in SF(m)$ with $f(1/2) = y$ that

$$\|f - c_{ym}\| = \|f - c_{ym}\|_\infty \leq 1.$$

Thus for any $y \in [-1, 1]$,

$$\text{Rad}\{f \in SF_0 : f(1/2) = y \text{ and } |f(t) - y| < 1 \text{ for } |t - 1/2| \leq 1/m\} \leq 1$$

so by (1.16), $R(N, SF(m)) \leq 1$.

It is not hard to show that the above inequality is an equality, so

$$R(N, SF(m)) = 1.$$

Now define $1 - \delta_m = \mu(F(m))$. Then $\delta_m \xrightarrow[m \to \infty]{} 0$, and since the probabilistic radius is

$$R_\delta^{\text{prob}}(N) \equiv \inf_{\mu(A) \leq 1-\delta} R(N, SA),$$

we have that for $m$ a positive integer,

$$R_{\delta_m}^{\text{prob}}(N) \leq R(N, SF(m)) = 1.$$

Thus for all $\delta > 0$,

$$R_\delta^{\text{wor}}(N) < 1.$$

Thus we have an example where $R^{\text{wor}}(N) = 2$, whereas $R_\delta^{\text{prob}}(N) \leq 1$.

Now taking infima over $N$ to get minimal radii of information, we have by the above

$$R^{\text{wor}}(1) = \inf_{\text{card } N=1} R^{\text{wor}}(N) = 2,$$

while

$$R_\delta^{\text{prob}}(1) = \inf_{\text{card } N=1} R_\delta^{\text{prob}}(N) \leq 1,$$

showing that for arbitrarily small $\delta$, probabilistic radii of information (which represent lower bounds for error) can be quite different from worst case radii.

The intuition here for the case $Nf = f(1/2)$ is the following. In our error metric, suppose we want to estimate a Brownian path $f(t)$ from knowing $f(1/2)$. Then if we are willing to ignore a set of arbitrarily small probability (essentially ignoring functions $f(t)$ which approach their values $f(1/2)$ very suddenly), we can cut error in half, no matter how small a set we throw away. This is somewhat unexpected. In this case it does not pay to heed the worst case scenario, since functions whose possibility we are ignoring can have arbitrarily small probabilities.

This example can be extended to information operators $N$ of higher cardinality, giving values of the Brownian path at several points, as opposed to just one.

In addition, though the above example for Brownian motion involves a space F which is non-separable, this is also not essential. One can replace the Wiener space $F$ by the following restriction $X$ of Wiener space to a discrete sequence space. Consider the set $A$ of real-valued functions $f(1/2 - 1/n)$, defined only on $a_n = 1/2 - 1/n$, for $n = 1, 2, 3, \ldots$, with the property that $\lim_{n \to \infty} f(1/2 - 1/n)$ exists. Endow $A$ with the supremum norm, making it a Banach space. Note this space is separable, being the space of all real-valued sequences with limits. Now let us give $A$ the following Gaussian measure. First define a map $M$ from $F$ (Wiener space) into $A$ by $M(f) = \{f(1/2 - 1/n)\}_n$ (thus $M$ takes $f$ to the sequence defined by its restriction on the points $\{1/2 - 1/n\}$. It is easy to check that the function $f(1/2 - 1/n)$ is indeed in $A$. Note also the map $M$ is bounded with norm 1. Define the Gaussian measure $\nu$ on $A$ by defining, for any Borel set $B \subset A$,

$$\nu(B) = \mu(M^{-1}(B)).$$

Since $M$ is continuous, this defines a Borel measure on $A$, and it is easy to check that it is Gaussian (note that applied to all cylinder sets the measure is clearly Gaussian).

Using the arguments above, if we let the space $B$ be $A$ endowed with the norm $\|f\| = \|f\|_{L^\infty} + |f(1/2)|$ (where $f(1/2) \equiv \lim_{n \to \infty} f(1/2 - 1/n)$), and define the linear operator $S : A \to B$ by $Sf = f$, it can be shown in the same way as above that for the fixed information operator $Nf = f(1/2)$, $R_\delta^{\text{prob}}(N) \leq 1$ for all $\delta > 0$, while for any linear functional $N$, $R^{\text{wor}}(N) = 2$, so that earlier comparison between worst case and probabilistic radii of information holds here.

The above can be summarized in:

**Theorem 4:** *Let $\mathcal{N}$ be all continuous linear functionals or just linear functionals consisting of standard information, i.e., pointwise evaluation. Then for S the identity operator from F to*

*G above (in either the Wiener space or its discrete restriction), the worst case radius $R^{wor}$(evaluated as an infimum over $\mathcal{N}$) is strictly larger, by a factor of at least 2, than $\lim_{\delta \to 0} R_\delta^{prob}$. The same is true if we evaluate the above radii of information for the fixed information operator $Nf = f(1/2)$.*

*Thus Theorems 2 and 3 are false if uniform convexity for G is not assumed.*

**Remarks: 1**. This type of result of course extends to situations where the information operators $N$ have arbitrary cardinality n, with an appropriate modification of the norm on $G$.
**2.** This theorem shows there are scenarios in which total error can be cut in half (below that in the worst case setting) in a completely risk-free way, by ignoring a set of problem elements whose total probability is arbitrarily small.

We also remark that though one might suspect on the basis of Theorem 3 that the probabilistic limit is the same as the worst case limit for a much more general class of measures than the nonvanishing ones defined here (see Def. 2.2) this is definitely not the case, even for the class of measures which (along with their conditionals and marginals) are nonvanishing on all open sets. To see this, consider $F = \mathbb{R}^2$ with Euclidean norm, and the measure $\mu = \nu \times \lambda$ restricted to the unit ball $F_0$ and normalized, where $\lambda$ is Lebesgue measure and $\nu$ is the point measure $\sum_i c_i \delta_{x_i}$ with $\sum_i c_i = 1$, and $\{x_i\}_{i=1}^\infty$ an enumeration of the rationals in $\mathbb{R}$. It can be shown that the marginals and conditionals of $\nu$ can be chosen in such a way that they do not vanish on any open set. However, it can also be shown that the identification problem with $S: F_0 \to F_0$ the identity has $R_1^{wor} = 1$, where $R_1^{wor}$ is the minimal worst case radius of information for information of cardinality 1, while $R_{\delta,1}^{prob}=0$ for all $\delta > 0$, where $R_{\delta,1}^{prob}$ denotes the minimal $\delta$-probabilistic radius for information of cardinality 1.

As corollaries of the results used for analyzing the worst case limit of probabilistic recovery, there are others of basic interest in the geometry of Banach space and in its applications to continuous complexity. Though some of these are presumably known in the theory of Banach spaces, they are listed here for completeness.

**Proposition 5:** *Let $F$ and $G$ be Banach spaces and let $N : F \to \mathbb{R}^n$ be linear. Let $S : F \to G$ be a continuous linear operator. Then for $B \subseteq F$ a closed convex set, the function $R(N,y) \equiv \text{Rad } S(N^{-1}(y) \cap B)$ is continuous in $(N,y)$, taken in the uniform operator topology crossed with the topology of $\mathbb{R}^n$, in the interior $I$ of the support of the function $R$, if the operators $N$ are restricted to have given fixed rank $n$*

**Proposition 6:** *Every set in a uniformly convex Banach space has a unique center.*

**Proposition 7:** *Let $F$ be a Banach space, and $G$ be a uniformly convex Banach space. Given arbitrary (adaptive or nonadaptive) information $N$, a strongly optimal algorithm for approximating $S : F \to G$ always exists and is unique.*

**Proposition 8:** *In a uniformly convex Banach space the radius functional is continuous in the*

*topology of set convergence, in that if* $A_n \overset{\uparrow}{\underset{n\to\infty}{}} A$, *then* $\text{Rad}(A_n) \underset{n\to\infty}{\longrightarrow} \text{Rad}(A)$.

**Proposition 9:** *In any separable uniformly convex Banach space $F$, the radius of a set $E$ is the supremum of the radii of its finite subsets,*

$$\text{Rad}(E) = \sup_{C \subseteq E, C \text{ finite}} \text{Rad}(C).$$

**Proposition 10:** *For a fixed adaptive or nonadaptive information operator $N$ and a convex set $A$ in a Banach space $F$, the $N$-radius $R(N, A)$ of $A$ is supported on finite dimensional subsets. That is, for any set $A$ there exist finite dimensional subsets $A_n \subseteq A$ such that $R(N, A_n) \underset{n\to\infty}{\longrightarrow} R(N, A)$. The same is true for partially defined information operators $N$.*

We also prove a result on the relationship of alternative definitions of average complexity (using $L^p$ norms instead of $L^2$ norms), to worst case complexity.

**Theorem 11:** *Assume $\mu$ is a nonvanishing measure supported on the unit ball $F_0 \subset F$, with $F$ separable Banach space, and $S : F \to G$, is a bounded linear operator, with $G$ a uniformly convex Banach space. Define the p-average local radius of information analogously to the average radius, but using an $L^p$ norm,*

$$R_p^{\text{avg}}(N) = \inf_{h:\mathbb{R}^n \to F} \left( \int_F \|Sf - Sh(Nf)\|^p \, d\mu(f) \right)^{1/p}.$$

*Then as $p \to \infty$,*

$$R_p^{\text{avg}}(N) \underset{p\to\infty}{\longrightarrow} R^{\text{wor}}(N).$$

*Further, if we take infima over $N$ of cardinality $n$,*

$$\lim_{n\to\infty} R_{p,n}^{\text{avg}} = R_n^{\text{wor}} \equiv \inf_{\text{rank } N = n} R^{\text{wor}}(N). \tag{1.5}$$

*In addition, for the corresponding $\epsilon$-cardinalities of information*

$$\text{card}_p^{\text{avg}}(\epsilon) \underset{p\to\infty}{\longrightarrow} \text{card}^{\text{wor}}(\epsilon),$$

*for every $\epsilon$ at which* $\text{card}^{\text{wor}}$ *is continuous.*

Above, the infima over operators $N$ are taken over the class with *worst case* cardinality $n$, as opposed to average cardinality. This is because it does not make sense to take infima over information operators $N$ of average cardinality $N$ in the definition of $R_p^{\text{avg}}(n)$ as $p \to \infty$, since in the worst case (i.e., essentially $p = \infty$) setting, $n$-radius of information is defined as an infimum over the class of operators $N$ of worst case cardinality $n$, as opposed to average cardinality $n$. The statement of the theorem changes if we define the $p$-average radius of information of order $n$ by taking infima over $N$ of *average* cardinality $n$, to say that generically the convergence in (1.5) does not occur, since then different definitions of cardinality would be used on the right and left sides.

We define the *adaptive Gelfand $n$-radius* of a set $A$ in a Banach space $F$ as a natural generalization of the Gelfand $n$-width, more appropriate to the notion of adaptive estimation. Namely, let the adaptive Gelfand radius (or just Gelfand radius) of order $n$ of the set $A$ be defined as

$$R_n^{\text{wor}}(A) = \inf_{N:F\to\mathbb{R}^n,\, N \text{ adaptive}} \sup_{y\in\mathbb{R}^n} \text{Rad}(N^{-1}(y) \cap A).$$

As opposed to the standard notion of a Gelfand $n$-width, this measures a radius minimized over the intersections of the set $A$ with larger classes of subspaces.

This notion of radius, when applied to a set $F_0$ of problem elements in a normed linear space $F$, gives exactly the adaptive radius of information. It is interesting to ask how this radius of information changes with varying *a priori* information. Assume that there is a nonadaptive information operator $M$, and we are given *a priori* information that our problem element $f \in F_0$ satisfies $Mf = z$. In this case, what is the smallest error (radius of information) in approximating $f$ using $n$-adaptive information? More specifically, does this error vary continuously with $M$ and $z$? We conjecture it does, and prove a partial result (lower semicontinuity) on the minimum error in this situation (Theorem 11).

Another important motivation for this result has to do with the fact that the adaptive Gelfand radius can be defined recursively (assuming the adaptive information $N$ is given as $(L_1, L_2, \ldots, L_n)$) by:

$$R_n^{\text{wor}}(A) = \inf_{N:F\to\mathbb{R}^d;\, N \text{ adaptive}} \sup_y \text{Rad}(N^{-1}(y) \cap A)$$

$$= \inf_{L:F\to\mathbb{R};\, L \text{ linear}} \sup_y R_{n-1}(L^{-1}(y) \cap A).$$

It is sometimes useful to know that in this recursive definition, $R_{n-1}^{\text{wor}}(L^{-1}(y) \cap A)$ is lower semicontinuous in $y$ and $L$. We have the following theorem.

**Theorem 12:** *Let F and G be Banach spaces with F separable and G uniformly convex, and let N: $F \to Y = \mathbb{R}^n$ be nonadaptive information operators assumed to remain of fixed rank. Let $S : F \to G$ be a continuous linear operator. Then the function $R(N, y, B) \equiv \text{Rad } S(N^{-1}(y) \cap B)$ is lower semicontinuous in $(N, y)$, in the interior $I$ of the support of the function $R$. Here we assume the topology of uniform convergence on $\mathcal{N}$, i.e., that $N_k \xrightarrow[n\to\infty]{} N$ if for each $i$ and each choice $y_1, \ldots, y_{i-1}$, $L_{ki}(y_1, \ldots, y_{i-1}) \xrightarrow[k\to\infty]{} L_i(y_1, y_2, \ldots, y_{i-1})$ in the uniform topology on linear functionals, where $L_{ki}$ is the $i^{th}$ component of $N_k$.*

Finally, we consider computational complexity, by taking algorithms into account. Results in [TWW, section 8.5.4] on optimality of spline algorithms do not apply here, since we are not dealing with Gaussian measures, nor with the entire space $F$, but a measure $\mu$ restricted to a bounded convex subset $F_0$.

Nevertheless, we show that under the most general model of combinatory cost for the algorithm $\phi$, we can talk not only about the approach of the cardinality of information to the worst case limit as $\delta \to 0$, but also regarding full computational complexity. We will assume that the combinatory cost denoted by $\text{cost}(\phi, N(f))$ of the optimal algorithms $\phi$ may have any dependence on the operator $N$ and information $y = Nf$. Our theorem applies to any situation in which uniformly continuous families of optimal (i.e., optimal on a set of probability $1 - \delta$) probabilistic algorithms exist. This occurs, for example, if we restrict to linear algorithms of some uniform bound, to any class of uniformly continuous $\phi$, or in any situation where we can prove uniformly continuous families of $\delta$-optimal algorithms exist. The theorem shows also that if the hypothesis of some type of continuity is taken away, the statement that probabilistic complexity approaches worst case complexity is false.

Throughout our discussion we assume we are dealing with an allowed class $\Phi$ of algorithms (e.g., all algorithms, all continuous algorithms, all linear algorithms, etc.). All infima over $\phi$ will be assumed to be over this class. The probabilistic computational complexity has the form

$$\text{comp}_\delta(\epsilon) =$$

$$\inf_{\Phi;N} \{c \text{ card } N + \sup_{f \in F_0} \text{cost}(\phi, N(f)) : \inf_{\mu(A_\delta) \geq 1-\delta} \sup_{f \in A_\delta} \|\phi(N(f)) - Sf\| \leq \epsilon\},$$

where $c$ denotes the cost of each information operation, and $\text{cost}(\phi, N(f))$ denotes the so-called combinatory cost of using information $N(f)$ and the algorithm $\phi$ to arrive at an approximation $\phi(N(f))$. Above and henceforth it is assumed that the infimum over $A_\delta$ is taken only over $A_\delta \subseteq F_0$. Worst case complexity $\text{comp}^{\text{wor}}(\epsilon)$ has the same form as above, with $A_\delta$ replaced by $F_0$ and the infimum over $A_\delta$ eliminated.

Define

$$\text{cost}(\phi, N) = \sup_{f \in F_0} \text{cost}(\phi, N(f)),$$

$$\text{comp}(\phi, N) = c \text{ card } N + \text{cost}(\phi, N) \tag{1.6}$$

(1.28)

and

$$e_\delta(\phi, N) = \inf_{\mu(A_\delta) \geq 1-\delta} \sup_{f \in A_\delta} \|\phi(N(f)) - Sf\|.$$

We will prove:

**Theorem 13:** (i) *Let $F$ and $G$ be Banach spaces, $S : F \to G$ be a continuous operator, and $\mu$ be a nonvanishing measure on the unit ball $F_0$ of $F$. Assume the worst case $\epsilon$-complexity is bounded for all $\epsilon$. Then if for given $\epsilon$ and arbitrarily small $\delta$ there exist information* *operators $N_\delta$ and algorithms $\phi_\delta \in \Phi$ such that* $\text{comp}(\phi_\delta, N_\delta) - \text{comp}_\delta(\epsilon) \xrightarrow[\delta \to 0]{} 0$ *and the $\phi_\delta(y)$ are continuous in $y$, uniformly in $\delta$, it follows that the probabilistic complexity approaches worst case complexity, i.e.,*

$$\text{comp}_\delta(\epsilon) \xrightarrow[\delta \to 0]{} \text{comp}^{\text{wor}}(\epsilon).$$

(ii) *if either of the assumptions regarding bounded complexity or continuity of the $\phi$ is removed, assertion (i) becomes false.*

## 2. Background and Definitions

Throughout this paper, we refer to terminology and formalism more fully described in [TWW]. Hilbert and Banach spaces will be real unless otherwise specified. Our theorems will generally deal with classes of nonvanishing measures on a Banach space (see definition below). This includes Gaussian and more general classes. A more general collection of measures on a Banach space is the so−called *orthogonally invariant measures*. In infinite dimension, this is the class of measures $\mu$ with

$$\mu(B) = \int_0^\infty g\left(\frac{1}{\sqrt{t}}\right) d\alpha(t) \tag{2.1}$$

for every Borel set $B \subseteq F$, where $g$ is a fixed Gaussian measure on $F$, and $\alpha$ a measure on $\mathbb{R}^+$ such that

$$\int_0^\infty d\alpha(t) = \int_0^\infty t \, d\alpha(t) = 1.$$

In infinite dimension, such measures coincide with the so−called *elliptically contoured measures*. In finite dimension, orthogonally invariant measures can be defined as measures with constant densities on scaled families of ellipses, i.e., measures $\mu$ with densities $w(\|Mx\|)$, where $M$ is s positive definite linear transformation, and $w: \mathbb{R}^+ \to \mathbb{R}^+$ is properly normalized.

We now give the essentials of probabilistic analytic complexity (see [W]).

**Definition 2.1:** Let $F$ be a linear space. An linear operator $N: F \to \mathbb{R}^n$ is called a *nonadaptive* information operator. An operator $N: F \to \mathbb{R}^n$ is called an *adaptive* information operator if

$$NF = (L_1 f, L_2(y_1)f, L_3(y_1, y_2)f, \ldots, L_n(y_1, \ldots, y_{n-1})f),$$

where each $L_i(y_1, \ldots, y_{i-1})$ is a bounded linear functional (by convention of norm 1) which is allowed to depend in any way on $y_1, \ldots, y_{i-1}$, where by definition

$$y_i \equiv L_i f.$$

**Definition 2.2:** Let $F$ be a Banach space with unit ball $B$ and a probability measure $\mu$ on $B$. Let $S$ be a linear map from $F$ into a Banach space $G$. Let $N : F \to \mathbb{R}^n$ be an information operator. Let $0 \leq \delta \leq 1$. The probabilistic radius $R_\delta^{\text{prob}}(N)$ is defined by

$$R_\delta^{\text{prob}}(N) \equiv \inf_{A \subset F_0, \mu(A) \geq 1-\delta} R(N, SA), \text{ 2.2} \qquad (2.2)$$

where $R(N, SA)$ is the radius of information of the operator $S$ with information $N$ on the admissible set $A$, i.e.,

$$R(N, SA) = \sup_{y \in \mathbb{R}^n} \text{Rad}(S(N^{-1}(y) \cap A)).$$

We should note that the Definition 2.2 is different from the form in which it appears elsewhere, e.g., in [TWW]. There the definition is made as follows. First for a given function (algorithm) $\phi : \mathbb{R}^d \to G$, we define the probabilistic error $e^{\text{prob}}(\phi, N, \delta)$ by:

$$e^{\text{prob}}(\phi, N, \delta) = \inf_{A; \mu(A) \geq 1-\delta} \sup_{f \in A} \|S(f) - \phi(Nf)\|,$$

where again all sets $A$ in the infimum are understood to be contained in $F_0$. Then the probabilistic radius there is defined by

$$R_\delta^{\text{prob}}(N) = \inf_\phi e^{\text{prob}}(\phi, N, \delta).$$

However, we see that therefore

$$R_\delta^{\text{prob}}(N) = \inf_{A: \mu(A) \geq 1-\delta} \inf_\phi \sup_{f \in A} \|S(f) - \phi(N(f))\|$$

$$= \inf_{A: \mu(A) \geq 1-\delta} \inf_\phi \sup_{y \in \mathbb{R}^d} \sup_{f \in N^{-1}(y) \cap A} \|S(f) - \phi(N(f))\|$$

$$= \inf_{A: \mu(A) \geq 1-\delta} \sup_{y \in \mathbb{R}^d} \inf_{g \in G} \sup_{f \in N^{-1}(y) \cap A} \|S(f) - g\|$$

$$= \inf_{A: \mu(A) \geq 1-\delta} \sup_{y \in \mathbb{R}^d} \text{Rad}(S(N^{-1}(y) \cap A))$$

$$= \inf_{A: \mu(A) \geq 1-\delta} R(N, SA),$$

as in (2.2).

**Definition 2.3:** Let $F$ be a Banach space and $\mu$ a finite Borel measure on $B \subseteq F$. We call $\mu$ *nonvanishing* if for every finite dimensional subspace $U$ of $F$, the conditional measure $\mu_U$ of $\mu$ restricted to $U$ satisfies $\mu_U(P) > 0$ for every $P \subseteq U \cap B$ with positive Lebesgue measure, and the same holds for the marginal measure on any finite dimensional subspace U with respect to a complementary subspace. Thus Lebesgue measure is absolutely continuous with respect to $\mu$ on finite dimensional restrictions.

If $\mu$ is a Borel measure on $B$ and $\mu_1$ is a marginal measure of $\mu$, the corresponding conditional measures $\mu(\,\cdot\,|y)$ are defined only up to $y$ in sets of $\mu_1$-measure 0; thus we define two sets of conditionals $\mu(\,\cdot\,|y)$ and $\mu/(\,\cdot\,|y)$ to be equivalent if they differ only on a set of $y$-$y$-measure 0. Thus, more precisely, we say a measure $\mu$ is nonvanishing if for every marginal measure $\mu_1$, the set of equivalent families of conditional measures has a member $\{\mu(\,\cdot\,|y)\}_y$ such that Lebesgue measure is absolutely continuous with respect to it for all $y$.

**Definition 2.4:** Let $F$ be a probability space with measure $\mu_0$, and $B \subseteq F$, with $\mu(B) > 0$. Then the *restriction* $\mu = \mu_0|_B$ is the measure concentrated on $B$ (with domain $\{E \cap B : E \text{ is } \mu_0\text{-measurable}\}$) defined by

$$\mu(E) = \frac{\mu_0(E)}{\mu_0(B)}$$

for $E \subseteq B$ and $E$ $\mu_0$-measurable. .

We now prove Proposition 5 on the continuity of the radius of information. We require the following definitions and lemma:

**Definitions 2.5:** We will let $d$ denote the distance in the metric induced by the norm of $F$. We will also denote $d$ as the distance in $G$ if there is no confusion. Let $I$ denote the interior of the set $\{(N, y) : N \text{ has fixed rank } k \text{ and } RD(N, y) \neq 0\}$. As usual $B_\epsilon(x)$ denotes the closed ball of radius $\epsilon$ about the point $x$.

**Definition 2.6:** Let $d$ be the distance in a metric space $M$, and assume the usual notion of distance $d(g, V) = \inf_{f \in V} d(g, f)$ between a point and a set. Let $V_1, V_2, \subseteq M$. Denote the maximal set distance between $V_1$ and $V_2$ by $\delta(V_1, V_2) = \max\left(\sup_{g \in V_1} d(g, V_2),\ \sup_{g \in V_2} d(g, V_1)\right)$.

**Lemma 1:** If $W_n$ and $W$ are sets in a Banach space $X$ and $\delta(W_n, W) \xrightarrow[n \to \infty]{} 0$, then $\text{Rad}(W_n) \xrightarrow[n \to \infty]{} \text{Rad}(W)$.

*Proof*: We have $\text{Rad}(W_n) \equiv \inf_{f \in X} \sup_{g \in W_n} \|f - g\|$. Further, for any $\epsilon > 0$,

$$\text{Rad}(W) - \epsilon = \inf_{f \in X} \left( \sup_{g \in W} \|f - g\| - \epsilon \right)$$

$$\leq \inf_{f \in X} \sup_{g \in W_n} \|f - g\| = \text{Rad}(W_n)$$

$$\leq \inf_{f \in X} \left( \sup_{g \in W} \|f - g\| + \epsilon \right) = \text{Rad}(W) + \epsilon$$

for n sufficiently large. Hence $\text{Rad}(W_n) \xrightarrow[n \to \infty]{} \text{Rad}(W)$. $\xrightarrow[n \to \infty]{}$. $\square$

*Proof of Proposition 5*: Let $(N_n, y_n) \to (N, y) \in I$ in our topology. We will show that $R(N_n, y_n) \xrightarrow[n \to \infty]{} R(N, y)$, and hence the radius of information is continuous. Let $W = N^{-1}(y) \cap B$ and $W_n = N_n^{-1}(y_n) \cap B$. Our approach is to show that under the above assumption $\delta(W_n, W) \xrightarrow[n \to \infty]{} 0$, and hence the same is true of $\delta(SW_n, SW)$, so that $\text{Rad}(SW_n) \xrightarrow[n \to \infty]{} \text{Rad}(SW)$ by the sub-lemma, showing the radius of information is continuous.

We show that $\delta(W_n, W) \xrightarrow[n \to \infty]{} 0$ by contradiction as follows. irst, for any sets $V_1$ and $V_2$, if $\delta(V_1, V_2 > 2\epsilon$, either

$$\sup_{f \in V_1} d(f, V_2) \geq 2\epsilon \quad \text{or} \quad \sup_{f \in V_2} d(f, V_1) \geq 2\epsilon.$$

In the first case we have $B_\epsilon(f) \cap V_2 = \emptyset$, for some $f \in V_1$, and in the second $B_\epsilon(f) \cap V_1 = \emptyset$ for some $f \in V_2$.

Thus assume for a contradiction that $\delta(W_n, W)$ does not go to 0, (by taking subsequences we will assume that $\delta(W_n, W) > 2\epsilon$ for all $n$, for some $\epsilon > 0$). Hence at least one of the following will hold:

CASE I: $B_\epsilon(f_{n_i}) \cap W_{n_i} = \emptyset$ for some subsequence $\{n_i\}$, with $f_{n_i} \in W_{n_i}$.

CASE II: $B_\epsilon(f_{n_i}) \cap W_{n_i} = \emptyset$ for some subsequence $\{n_i\}$, with $f_{n_i} \in W$.

We will find a contradiction in either case by showing that it follows that $(N, y) \notin I$. To this end, we claim it suffices to show $W = (N^{-1}(y) \cap B) \subseteq \partial B$.

To verify this, assume $W \, \partial B$. Note that the interior $B^o$ of $B$ is convex (since $B$ is), and that $N^{-1}(y) \cap B^o = \emptyset$. Thus $y \notin N(B^o)$. In addition $N(B^o)$ is convex since $N$ is linear. By the Hahn-Banach theorem there exists a sequence $\{z_i\} \subset \mathbb{R}^n$ such that $z_i \to y$, and $d^*(z_i, N(B^o)) > 0$ for all i, where $d^*$ denotes Euclidean distance. Hence $d(N^{-1}(z_i), B^o) > 0$ for all $i$. Hence $N^{-1}(z_i) \cap B = \emptyset$, and so $R(N, z_i) = 0$ for all $i$. Since $z_i \to y$, the point $(N, y)$ is on the boundary of the set where $R(N, y)$ is nonvanishing, so it is not in its interior, proving that $(N, y) \notin I$, as desired.

Thus it is left to prove that in either Case I or Cse II, $W \subset \partial B$.

We first show that for any $g \in W$, $d(g, N_n^{-1}(y_n)) \xrightarrow[n \to \infty]{} 0$, uniformly in $g$. Thus let $g \in W$. Let $X_1$ be a subspace complementing Ker $N$ in $F$. Consider $N$ as a linear operator, restricted to the affine subspace $X_2 \equiv g + X_1$. Since $N|_{X_2}$ is bijective and $X_2$ is finite dimensional, there is a constant $C$ such that for $h \in X_2$.

$$\|N(h)\| > C\|h\|_{X_2}, \qquad 2.3$$

where $\|h\|_{X_2}$ denotes the norm relative to $X_2$, i.e., $\|h\|_{X_2} = \|h - g\|_X$. Thus

$$\|N(h)\| > C \|h - g\|$$

for $h \in X_2$. Consider the affine map $N_n|_{X_2}$. For $n$ sufficiently large,

$$\|N_n|_{X_2} - N|_{X_2}\| \equiv \sup_{x \in X_2;\, \|x\|_{X_2} \leq 1} \|N_n x - N x\|$$

can be made arbitrarily small (uniformly in $g \in W = \text{Ker } N \cap B$, since all $x$ which can occur can in the supremum are from a set in $F$ which is uniformly bounded). Thus there exists a constant $M$ such that for $n > M$, $N_n|_{X_2}$ is a bijective affine map, i.e., an affine map whose gradient never vanishes. Note further that the choice of $M$ can be made uniform in $g \in W$, because the gradient of $N\|_{X_2}$ is bounded away from $0$, uniformly in $g \in W$. Let $N_n|_{X_2} f \equiv N_n^L f + \nu_n$ be the decomposition of $N_n|_{X_2}$ into a linear map and a translation. Then if

$$\|N_n|_{X_2} - N|_{X_2}\| \equiv \sup_{\|h\|_{X_2} \leq 1} \|N_n h - N h\| < \epsilon,$$

$$\epsilon > \|N_n|_{X_2}(0) - N|_{X_2}(0)\| = \|\nu_n\|_{X_2},$$

so that $\|\nu_n\|_{X_2} < \epsilon$. Note therefore that as $n \to \infty$, $N_n^L \to N|_{X_2}$ in the norm topology, uniformly in $g$ as above. Consequently, there exists a $c > 0$ and an $M$ (possible increased from the $M$ above) such that if $n > M$, then $\|N_n^L h\|_{X_2} \geq c\|h\|_{X_2}$ for $h \in X_2$. Again note that this estimate can be made uniformly in $g \in W$. Let $z_n \in X_2$ be the zero of $N_n|_{X_2}$. Then $N_n^L z_n + \nu_n = 0$. Hence

$$\|v_n\|_{X_2} = \|N_n^L z_n\|_{X_2} \geq c\|z_n\|$$

for $n > M$. Thus for such $n$, $\|z_n\|_{X_2} \leq (1/c)\|v_n\| \to 0$, uniformly in $g$. Thus if $\delta > 0$, there exists $M > 0$ such that if $n > M$, then for all $g \in W$, there exists a $z \in \text{Ker } N_n$ such that $\|z - g\|_X < \delta$. Equivalently,

$$d(g, \text{Ker } N_n) < \delta. \qquad 2.3$$

uniformly in $g \in \text{Ker } N$, for $n$ sufficiently large.

Similarly, it can be shown that $d(h, \text{Ker } N) \leq \delta$ uniformly in $h \in W_n$, for $n$ sufficiently large. Indeed, choose $g \in \text{Ker } N$ so that $h \in X_2 \equiv g + X_1$ as above. Then for $C$ as in (2.3), $\|Nh\| \leq Cb$ for $n$ sufficiently large, uniformly in $h$, since $N_n h = 0$ and $N_n \xrightarrow[n \to \infty]{} N$ in the uniform operator topology. Since we also have $C\|h\|_{X_2} \leq \|Nh\|$, we conclude that $\|h\|_{X_2} \leq \delta$, i.e., $d(h, \text{Ker } N) \leq \delta$, again uniformly in $h \in W_n$.

Now assume first that CASE I above holds (for a contradiction). Thus there is a subsequence $\{n_i\}_i$ and points $f_{n_i} \in W_{n_i}$ for which $d(f_{n_i}, W) > \epsilon$ for some fixed $\epsilon > 0$. Again by taking subsequences, assume this is so for all $n$. By the result of the previous paragraph, for any $\delta > 0$, and for $n$ sufficiently large, $d(h, N^{-1}(y)) < \delta$, uniformly for $h \in W_n$. Hence if $h \in B_\epsilon(f_n) \cap W_n$, then $d(h, N^{-1}(y)) < \delta$. Recall of course that $B_\epsilon(f_n) \cap W = \emptyset$ under the present assumption of CASE I. We will show from this that $W \subseteq \partial B$. To this end, let $f \in W$. Choose $\delta$ and $n$ as above. Consider the line segment $L$ from $f$ to $f_n$. Let $|L|$ be its length.

We claim that

$$d(f, \partial B) \leq 2(\delta/\epsilon)|L| \leq 4(\delta/\epsilon),$$

the latter inequality following from the fact that $|L| \leq 2$, $L$ being entirely in the unit ball $B$. To see this, assume it is false for a contradiction. Then we would have $B_{2(\delta/\epsilon)|L|}(f) \subset B$. Let $\mathcal{C}$ be the convex hull of $\{B_{2(\delta/\epsilon)|L|}(f) \cup \{f_n\}\}_n$. Under our assumption, $\mathcal{C} \subset B$. Let

$$h = ((\epsilon/2)/|L|)f + (1 - (\epsilon/2)/|L|)f_n,$$

so that $h$ is on $L$, and $\epsilon/2$ units away from $f_n$.

We then would have $B_\delta(h) \subset \mathcal{C} \subset B$. To see this, note that if $p \in B_\delta(h)$, then

$$p^* \equiv (|L|/(\epsilon/2))(p - f_n) + f_n \in B_{2(\delta/\epsilon)|L|}(f).$$

Indeed,

$$
\begin{aligned}
|p^* - f| &= |(|L|/(\epsilon/2))(p - f_n) + f_n - f| \\
&= |(|L|/(\epsilon/2))(p - f_n) - (f - f_n)| \\
&= (|L|/(\epsilon/2))|(p - f_n) - (\epsilon/(2|L|))(f - f_n)| \\
&= (|L|/(\epsilon/2)) |(p - f_n) - (h - f_n)| \\
&= (|L|/(\epsilon/2)) |p - h| < 2(\delta/\epsilon)|L|.
\end{aligned}
$$

Thus under the above assumption $p^* \in B$. We also have

$$p = ((\epsilon/2)/|L|)\, p^* + (1 - (\epsilon/2)/|L|) f_n,$$

so that $p \in \mathcal{C} \subset B$ as desired.

However, $B_\delta(h) \subset B$ is impossible for sufficiently small $\delta$ for the following reason. Since $h$ is on the line $L$ from $f_n$ to $f$, and since $d(f_n, N^{-1}(y)) < \delta$, and $f \in N^{-1}(y)$, it follows that $d(h, N^{-1}(y)) < \delta$. Hence there is at least one point $g \in N^{-1}(y)$ such that $\|g - h\| < \delta$. For $\delta$ sufficiently small, $B_\delta(h) \subset B_\epsilon(f_n)$, so that $B_\delta(h) \cap W = \emptyset$. Hence $g \notin W$. Since $g \in N^{-1}(y)$ and $W = B \cap N^{-1}(y)$, we conclude that $g \notin B$. This shows that $B_\delta(h) \subset B$ is impossible, giving us the desired contradiction. Thus (2.5) is proved.

Since $\delta$ above is arbitrarily small, we conclude $f \in \partial B$, as desired.

Now assume CASE II above holds. Thus we can assume there exist an $\epsilon > 0$ and a subsequence $\{n_i\}_{i=1}^\infty$ with a sequence of points $f_{n_i} \in W$, such that $d(f_{n_i}, W_{n_i}) > \epsilon$. Taking subsequences, assume without loss that $d(f_n, W_n) > \epsilon$ for all $n$. Here we again show that $W \in \partial B$ as follows. Again for any $f \in W$ and $\delta > 0$, we claim $d(f, \partial B) \leq 4\delta/\epsilon$. To prove this, assume it is false. Choose $n$ so that for any $g \in B_\epsilon(f_n) \cap W$, we have $d(g, N_n^{-1}(y_n)) < \delta$. This can be done by (2.4). Let $L$ and $h$ be defined as above relative to $f$ and $f_n$. Then as before, by our assumption we have $B_{2(\delta/\epsilon)|L|}(f) \subset B_{4\delta/\epsilon} \subset B$. By the same argument as above, $B_\delta(h) \subset B$. Again let $\delta$ be sufficiently small but fixed. Then $B_\delta(h) \cap N_n^{-1}(y_n) \neq \emptyset$, since $h \in B_\epsilon(f_n) \cap W$. Thus $B_\delta(h) \cap W_n \neq \emptyset$ for $n$ sufficiently large. But we can choose $\delta$ so small that $B_\delta(h) \subset B_\epsilon(f_n)$, so by our assumption above, $B_\delta(h) \cap W_n = \emptyset$. This gives the desired contradiction. Since $\delta$ was arbitrary, we conclude $f \in \partial B$, as desired.

We have shown by contradiction in both CASES that $\delta(W_n, W) \xrightarrow[n \to \infty]{} 0$, so that by the above arguments $R(N_n, y_n) \xrightarrow[n \to \infty]{} R(N, y)$, which is what was needed. $\square$

One can actually show more than Proposition 4. Not only is the radius of information continuous in information operator $N$ and element $y$, but it is also a lower semicontinuous function of the information $N$ with infimum taken over $y$.

**Corollary 2.2:** *For adaptive or nonadaptive information operators, the radius of information*

$$\mathrm{Rad}^{\mathrm{wor}}(N) = \sup_{y \in \mathbb{R}^d} \mathrm{Rad}^{\mathrm{wor}}(N, y)$$

*is lower semicontinuous in the information operator $N$ in the interior of its support, as a function of information operators $N$ of constant rank. Here we assume the topology of uniform convergence on the family $\mathcal{N}$ of adaptive information operators, i.e., that $N_k \xrightarrow[k \to \infty]{} N$ if for each $i$ and each choice $y_1, \ldots, y_{i-1}$, $L_{ki}(y_1, \ldots, y_{i-1}) \xrightarrow[k \to \infty]{} L_i(y_1, y_2, \ldots, y_{i-1})$ in the dual space topology of $F^*$.*

*Proof:* Let $N$ be a fixed adaptive information operator, and let $N_k(y, \cdot) \xrightarrow[k \to \infty]{} N(y, \cdot)$. Here the adaptive operator $N(\cdot)$ is written in the form $N(y, \cdot)$ to make explicit its adaptive dependence on the information $y = N(f)$. Namely, $N(y, \cdot)$ is the adaptive information operator defined by

$$N(y, f) = (L_1, L_2(y, f), L_3(y, f), \ldots, L_n(y, f)),$$

where $L_i(y, \cdot)$ as a linear functional depends only on the first $i-1$ coordinates of y. Note that even though in the notation above it is assumed that $y = N(f)$ has been isolated in its dependence on the first component $y$ of its argument and so for fixed $y$, $N(y, f)$ is a nonadaptive information operator. Thus in particular for $y$ fixed, by Proposition 4,

$$\text{Rad } N_k^{-1}(y) = \text{Rad } N_k^{-1}(y, y) \xrightarrow[k \to \infty]{} \text{Rad } N^{-1}(y, y) = \text{Rad } N^{-1}(y)$$

where $N^{-1}(y, y)$ is the inverse image of $y$ under the linear operator $N(y, f)$ (if the first argument $y$ is considered fixed). Now

$$\liminf_{k \to \infty} \text{Rad}^{\text{wor}}(N_k) = \liminf_{k \to \infty} \sup_{y \in \mathbb{R}^d} \text{Rad}(N_k^{-1}(y) \cap F_0)$$

$$\geq \sup_{y \in \mathbb{R}^d} \lim_{k \to \infty} \text{Rad}(N_k^{-1}(y) \cap F_0)$$
$$= \sup_{y \in \mathbb{R}^d} \text{Rad}(N^{-1}(y) \cap F_0)$$
$$= \text{Rad}^{\text{wor}}(N),$$

where we have used Proposition 5 in the second to last equality. This proves that $\text{Rad}^{\text{wor}}(N)$ is lower semicontinuous. $\square$

**Remark:** Though this argument does not prove it, we expect that the radius of information $R_n(N)$ is a fully continuous function of $N$ in the uniform operator topology.

## 3. Radii of sets in Banach spaces

We also require some lemmas about radii of sets in Banach spaces:

**Lemma 3.1:** *Let $E$ be a set in the metric space $X$. Let $T_r$ denote the collection of radius $r$ centers of E, i.e., $T_r \equiv \{a : B_r(a) \supseteq E\}$. Then $T_r$ is closed.*

*Proof*: Let $a_n \xrightarrow[n \to \infty]{} a$ be a convergent sequence of points, with $a_n \in T_r$ for all $n$. Then $E \subseteq B_r(a_n)$ for all $n$, so that for any $e \in E$, $d(e, a_n) \leq r$. Since $a_n \xrightarrow[n \to \infty]{} a$ it follows that $d(e, a) \leq r$, so that $e \in B_r(a)$. Thus $E \subseteq B_r(a)$, so that $a \in T_r$. $\square$

**Definition 3.1:** Let $F$ be a Banach space. Then $F$ is *uniformly convex* if for every $\epsilon > 0$, there is a $\delta = g(\epsilon) > 0$ such that $\|x\| = \|y\| = 1$ and $\|(1/2)(x+y)\| > 1 - \delta$ imply that $\|x - y\| \leq \epsilon$. The function $g(\epsilon)$ is the *modulus of convexity*. A set $A$ in a Banach space is uniformly convex if the Chebyshev norm induced by $A$ is uniformly convex.

Essentially, a Banach space is uniformly convex if its unit ball is uniformly convex, the latter meaning that there is a certain minimal curvature at each point on the surface of the unit ball.

**Lemma 3.2:** *Let $F$ be a uniformly convex Banach space.*
*(a) Then every set in $B$ has a unique center.*
*(b) Let $A$ be a set of radius $r$, and let $C(A, \epsilon) = \{x \in F : B_{r+\epsilon}(x) \supseteq A\}$ be the collection of centers of balls of radius $r + \epsilon$ containing $A$. There is a universal function $f(r, \epsilon)$ (depending only on $F$) such that the diameter $D(C(A,\epsilon)) \leq f(r,\epsilon)$. Further, $f(r,\epsilon) \xrightarrow[\epsilon \to 0]{} 0$, uniformly on compact $r$-subsets.*

*Proof*: We first prove (b), since (a) follows from it. Let $A \subseteq F$, and $C(A, \epsilon)$ be as above. We will give a universal upper bound on the radius $r \equiv \text{Rad}(A)$ as a function of $r$ itself and the diameter of the set of centers, $d \equiv D(C(A, \epsilon))$, thus getting a constraint on $D(C(A, \epsilon))$. Thus let $B_{r+\epsilon}(c_1)$ and $B_{r+\epsilon}(c_2)$ both contain $A$, with $\|c_1 - c_2\| = d - \delta$ (we will let $\delta \to 0$). Let $B = B_{r+\epsilon}(c_1) \cap B_{r+\epsilon}(c_2)$.

Note $A \subseteq B$. Further, $B$ is balanced with center $c = (c_1 + c_2)/2$. Indeed, if $b \in B_{r+\epsilon}(c_1)$, then its opposite about $c$ is $b' = c - (b - c) = 2c - b$. Note $\|b' - c_2\| = \|2c - b - c_2\| = \|(c_1 + c_2) - b - c_2\| = |c_1 - b| \leq r +$, so $b' \in B_{r+\epsilon}(c_2)$. Similarly if $b \in B_{r+\epsilon}(c_2)$ then $b' \in B_{r+\epsilon}(c_1)$. Hence if $b \in$ , so is $b'$, and $B$ is balanced about $c$. We now estimate the diameter of $B$, which will translate into an estimate on the radius since $B$ is balanced. Let $D(B)$ be the diameter. Let $P$ be a two dimensional subspace of $F$ containing $c_1$, and $c_2$ (and hence $c$). Let $t$ be the point of tangency to $P \cap B_{r+\epsilon}(c_2)$, of the parallel translation of the line containing $c_1 c_2$ so that it is tangent to both $P \cap B_{r+\epsilon}(c_1)$ and $P \cap B_{r+\epsilon}(c_2)$.

The problem is now in two dimensional geometry. Let $g(\epsilon)$ be the modulus of convexity of $F$. We can also consider the modulus of convexity $g_P(\epsilon)$ of $F \cap P$ in the plane $P$, defined by restriction of the norm in $F$ to $P$. It is easy to see from the definition that $F \cap P$ has the same or a larger modulus of convexity than $F$. Note that in finite dimension, $g_P(\epsilon)$ is invariant under all linear transformations, i.e., if we take the unit ball of a given norm and we transform it linearly, then it is the unit ball of a new norm, and the modulus of convexity of the new norm coincides with that of the old norm. Let $T$ be a one to one linear mapping of $P \cap F$ into a two dimensional coordinate system $\{(x, y) : x, y \in \mathbb{R}\}$. Assume $T$ has been chosen so that the image $T(c_1, c_2)$ of the line segment $c_1 c_2$ is orthogonal to $T(c_1 t)$ (in the usual sense in the $(x - y)$ plane). Further, assume the Euclidean diameters of $T(P \cap B_{r+\epsilon}(c_1))$ in the directions $T(c_1 c_2)$ and $T(c_1 t)$ both have length $2(r + \epsilon)$. For any $(x, y) \in \mathbb{R}^2$, define

$$\|(x,y)\| = \|T^{-1}(x,y)\|_F. \tag{3.1}$$

Then there is a $C$ such that for any $(x,y) \in \mathbb{R}^2$,

$$C^{-1}(x^2+y^2)^{1/2} \leq \|(x,y)\| \leq C(x^2+y^2)^{1/2}. \tag{3.2}$$

This compatibility allows us to define the modulus of convexity of the norm $\|(x,y)\|$ in terms of the Euclidean norm $\|(x,y)\|_2 \equiv (x^2+y^2)^{1/2}$. Namely, for $z_i = (x_i, y_i)$ with $\|z_i\| = 1$ ($i = 1,2$, if $\|z_1 - z_2\|_2 \geq \epsilon$, then $\|z_1 + z_2\|_2 \leq 1 - \delta$, where

$$\delta = \frac{1}{C} g(\epsilon/C)$$

this being obtained just by the comparison (3.2). This gives a "Euclidean" modulus of convexity for say the unit $\|\cdot\|$-ball of $\mathbb{R}^2$, and by scaling a Euclidean modulus of convexity may be obtained for any ball which is defined in $\mathbb{R}^2$ with respect to $\|\cdot\|$.

In this case, it is not difficult to show using two dimensional geometry that the $\|\cdot\|$-diameter of the intersection of the two balls $B_{r+\epsilon}(c_1) \cap P$ (in the Banach norm (3.2)) satisfies

$$D(P \cap B) = D(T(P \cap B)) \leq 2(r+\epsilon) h\left(\frac{d-\delta}{r+\epsilon}\right), \tag{3.3}$$

where $h(\cdot)$ is a function depending only on the modulus of convexity $g(\epsilon)$; the factor $2(r+\epsilon)$ represents the diameter of the original balls $B_{r+\epsilon}(c_i)$. The dependence of $h$ on the ratio $\frac{d-\delta}{r+\epsilon}$ reflects the fact that up to the overall scaling factor $2(r+\epsilon)$, the diameter (again in the Banach norm) of the intersection $T(P \cap B)$ is determined by ratio of the distance $d - \delta$ between the two balls, and their radius $r + \epsilon$ (all of which can be translated into compatible Euclidean distances). A way of justifying the two dimensional result (3.3) is to note that for any unit vector $v$ in $\mathbb{R}^2$, the diameter of $P \cap B$ in the direction $v$ is less than or equal to its value $2(r+\epsilon)$ when the separation $d - \delta$ of the two balls is 0, multiplied by a factor of at most $h\left(\frac{d-\delta}{r+\epsilon}\right)$, as can be seen from looking at the geometry in the above standardized representation in $\mathbb{R}^2$.

In addition, the function $h(x)$ satisfies $h(x) < 1$ for $x > 0$, and is decreasing. Since (3.3) holds for all planes $P$ (and since the maximal diameter of $B$ occurs on a line going through the center $c$ of $B$), it follows that

$$D(B) \leq 2(r+\epsilon) h\left(\frac{d-\delta}{r+\epsilon}\right).$$

Since $\mathrm{Rad}(B) = (1/2) D(B)$ (recall $B$ is balanced), (3.3) also gives an estimate for $\mathrm{Rad}(B)$, so that (since $A \subseteq B$) we have

$$r = \text{Rad}(A) \leq \text{Rad}(B) \leq (r+\epsilon)\, h\!\left(\frac{d-\delta}{r+\epsilon}\right)$$

Hence $h\!\left(\frac{d-\delta}{r+\epsilon}\right) \geq \frac{r}{r+\epsilon}$. This implies $\frac{d-\delta}{r+\epsilon} \leq h^{-1}\!\left(\frac{r}{r+\epsilon}\right)$, or

$$D(C(A,\epsilon)) = d \leq h^{-1}\!\left(\frac{1}{r+\epsilon}\right)(r+\epsilon) + \delta$$

Since this holds for all $\delta$, we may let $\delta \to 0$ to conclude

$$D(C(A,\epsilon)) \leq h^{-1}\!\left(\frac{r}{r+\epsilon}\right)(r+\epsilon) \equiv f(r,\epsilon).$$

Note that as $h^{-1}(1) = 0$, we have

$$f(r,\epsilon) \xrightarrow[\epsilon \to 0]{} 0,$$

and this convergence is uniform on compact $r$-subsets, which completes the proof of (b).

To show (a), let $A \subseteq F$. By Lemma 3.1, the collection $C(A, 1/n)$ of $r + 1/n$-centers of $A$ is closed, and $\text{Rad}(C(A, 1/n)) \leq D(C(A, 1/n)) \leq f(r, 1/n)$, for $n \in \mathbb{N}$. Let $c_n \in C(A, 1/n)$ for each $n$. Since the sets $C(A, 1/n)$ are closed and nested, i.e., $C(A, 1/n) \supseteq C(A, 1/(n+1))$, and since $\text{Rad}(C(A, 1/n)) \xrightarrow[n \to \infty]{} 0$ it follows the sequence $c_n$ is Cauchy, and that $c = \lim_{n \to \infty} c_n$ is in all of the sets $C(A, 1/n)$. Because $\text{Rad}(C(A, 1/n)) \xrightarrow[n \to \infty]{} 0$, it follows that $c$ is the only element of $\bigcap_n C(A, 1/n)$. Thus $A \subseteq B_r(c + 1/n)$, for all $n$ and so $A \subseteq B_r(c)$, so that $c$ is a center of $A$. The fact that $c$ is unique follows from the fact that it is unique in $\bigcap_n C(A, 1/n)$. $\square$

**Corollary 3.3:** *Let $F$ be a Banach space, and $G$ be a uniformly convex Banach space. Given an (adaptive or nonadaptive) information operator $N$, a strongly optimal algorithm for approximating $S : F \to G$ always exists and is unique.*

Indeed, a strongly optimal algorithm is a central algorithm, and the lemma shows that such an algorithm always exists and is unique in this case. We remark that such algorithms are generically nonlinear; see [KT1, 2].

**Definition 3.2:** Let $E$ be a set, and $A, A_n \subseteq E$ for $n = 1, 2, \ldots$ . Then we write $A_n \underset{n \to \infty}{\uparrow} A$ if $A_n \subseteq A_{n+1}$, and if $\bigcup_n A_n = A$.

**Proposition 8:** *In a uniformly convex Banach space the radius functional is continuous in the topology of set convergence, in the sense that if $A_n \underset{n \to \infty}{\uparrow} A$, then $\text{Rad}(A_n) \xrightarrow[n \to \infty]{} \text{Rad}(A)$.*

*Proof.* Assume that $A_n \uparrow_{n\to\infty} A$. Let $r = \sup_n \text{Rad}(A_n)$. For all $n$, let $T_{r+1/n}(A_n)$ denote the centers of all balls of radius $r+1/n$ which contain $A_n$. Then $T_{r+1/n}(A_n)$ is closed by Lemma 3.1. Since $A_n \subseteq A_{n+1}$, and $r+1/n \geq r+1/(n+1)$, we have $T_{r+1/n}(A_n) \supseteq T_{r+1/(n+1)}(A_{n+1})$. Thus the sequence $\{T_{r+1/n}(A_n)\}_n$ is a nested sequence of closed sets. Furthermore, by Lemma 3.2,

$$\text{Rad}(T_{r+1/n}(A_n)) \leq f(r_n, \epsilon_n),$$

where

$$r_n = \text{Rad}(A_n) \uparrow_{n\to\infty} r,$$

and

$$\epsilon_n = r + 1/n - r_n \downarrow_{n\to\infty} 0.$$

By Lemma 3.2, we have $f(r_n, \epsilon_n) \to_{n\to\infty} 0$, so

$$\text{Rad}(T_{r+1/n}(A_n)) \to_{n\to\infty} 0.$$

Hence by the same arguments as at the end of Lemma 3.2, there is unique point

$$a \in \bigcap_{n=1}^{\infty} T_{r+1/n}(A_n),$$

the intersection being of nested closed sets whose radii approach $0$. Hence for all $n$ and $\epsilon$ we have $A_n \subseteq B_{r+\epsilon}(a)$, so that

$$A = \bigcup_n A_n \subseteq B_{r+\epsilon}(a),$$

Hence $\text{Rad}(A) = r$, since any number smaller than $r$ would make the previous equation false for sufficiently small $\epsilon$. Thus $\text{Rad}(A_n) \to_{n\to\infty} \text{Rad}(A) = r$, as desired. $\square$

This lemma implies that in terms of radii, all sets can be approximated by finite sets in a separable uniformly convex Banach space. More precisely,

**Corollary 3.4:** *In any separable uniformly convex Banach space $F$, the radius of a set $E$ is the supremum of the radii of its finite subsets. That is,*

$$\text{Rad}(E) = \sup_{C \subseteq E, C \text{ finite}} \text{Rad}(C). \tag{3.4}$$

*Proof*: Let $E'$ be a countable dense subset of $E$, and $E'_n$ be a sequence of finite subsets such that $\text{Rad}(E'_n) \to_{n\to\infty} \text{Rad}(E')$ (this is possible by the lemma above). Then $\text{Rad}(E) = \text{Rad}(E') = \lim_{n\to\infty} \text{Rad}(E'_n)$. Since $\text{Rad}(E'_n) \leq \sup_{C \subseteq E, C \text{ finite}} \text{Rad}(C)$ (for all $n$,. we conclude

$$\operatorname{Rad}(E) \leq \sup_{C \subseteq E,\, C \text{ finite}} \operatorname{Rad}(C),$$

and the equality (3.4) follows. $\square$

We also have the following corollary, useful when the set $A$ whose radius we are measuring satisfies $\mu(A) = \mu(A^o)$, with $A^o$ the interior of $A$.

**Corollary 3.5:** *Let $F$ be a Banach space with nonvanishing measure $\mu$, and $A \subseteq F$. Let $G$ be a uniformly convex Banach space, and $S : F \to G$ be linear and bounded. If $A_n \subseteq A$ and $\mu(A_n) \xrightarrow[n \to \infty]{} \mu(A)$, then $\liminf_{n \to \infty} \operatorname{Rad}(SA_n) \geq \operatorname{Rad}(SA^o)$.*

*Proof.* Assume the conclusion is false. By taking subsequences we may assume that $\mu(A) - \mu(A_n) \leq 1 - 2^{-n}$. Let $B_n = \cap_{i \geq n} A_i$. Then $B_n \subseteq A_n$, $B_n \subseteq B_{n+1}$, and

$$\mu(B_n) \geq \mu(A) - \sum_{i \geq n} 2^{-i} = \mu(A) - 2^{-n+1} \xrightarrow[n \to \infty]{} \mu(A).$$

Let $B = \overset{\cup}{n} B_n \subseteq \overset{\cup}{n} A_n$. By the above, $\mu(B) = \mu(A)$, and $B \subseteq A$. Thus, $\mu(A^o) = \mu(B \cap A^o)$, so that $B \cap A^o$ is dense in $A^o$, since $\mu$ is nonvanishing. Hence $S(B \cap A^o)$ is dense in $SA^o$, so that $R(S(B \cap A^o)) = R(SA^o)$. Note that since $B_n \uparrow B$, we have $SB_n \uparrow SB$, so that $\operatorname{Rad}(SB_n) \uparrow \operatorname{Rad}(S(B))$. Hence $\operatorname{Rad}(SA_n) \geq \operatorname{Rad}(SB_n) \uparrow \operatorname{Rad}(SB)$. and $\liminf_{n \to \infty} \operatorname{Rad}(SA_n) \geq \operatorname{Rad}(SB)$. But $\operatorname{Rad}(SB) \geq \operatorname{Rad}(S(B \cap A^o)) = \operatorname{Rad}(SA^o)$. Combining these facts, we have $\liminf_{n \to \infty} \operatorname{Rad}(SA_n) \geq \operatorname{Rad}(SB) \geq \operatorname{Rad}(SA^o)$. Since this contradicts our assumption that the conclusion of the Corollary is false (which carried over to the present subsequence), we conclude that in fact $\operatorname{Rad}(SA_n) \xrightarrow[n \to \infty]{} \operatorname{Rad}(SA)$, as desired $\square$

We will also state a convergence lemma to be used in the theorem below.

**Lemma 3.6:** *Let $f_n(y)$ be a sequence of real-valued functions on a space $Y$, which are nondecreasing in $n$. Then*

$$\lim_{n \to \infty} \sup_y f_n(y) = \sup_y \lim_{n \to \infty} f_n(y).$$

In constructing the proof of Theorem 3, we will require the following simple lemma:

**Lemma 3.7:** *Let $F$ be a Banach space, and $L$ be a linear dense subset. Then $L$ is also dense in any closed subspace $K$ of $L$ which has finite codimension.*

*Proof*: We proceed by induction on codimension. Assume first that codim $K = 1$. Then $K$ separates $F$ into two disjoint components, $F_1$ and $F_2$. Let $x \in K$. Since $L$ is dense, there exists a sequence $\{x_k\} \subset L$ such that $x_k \xrightarrow[k \to \infty]{} x$, and such that $x_k \in F_1$; similarly there exists a sequence $\{y_k\} \subset L$ such that $y_k \xrightarrow[k \to \infty]{} x$, and such that $y_k \in F_2$ for all $k$. Thus for each $k$, there is a unique element $z_k$ contained in the line with endpoints $y_k$ and $x_k$ such that $z_k \in K$. It is easy to check that also that $z_k \xrightarrow[k \to \infty]{} x$; since $z_k \in L$, it follows from the fact that $x$ is arbitrary that $L \cap K$ is dense in $K$. Now assume that the lemma is true for all $K$ with codimension $n$. To show it for codimension $n+1$, let $K$ have codimension $n+1$. There is a subspace $K'$ of codimension $n$ containing $K$. Thus $L \cap K'$ is dense in $K'$ by the induction hypothesis, and $K$ has codimension 1 in $K'$. Thus the proof is completed by the application of the $n=1$ result. □

We will now require a notion involving *partially defined* information operators:

**Definition 3.2:** An (adaptive or nonadaptive) information operator $N(f)$ is *partially defined* if the component linear functionals $L_i(y_1, y_2, \ldots, y_{i-1})$ are not necessarily defined for all choices of $y_1, y_2, \ldots, y_{i-1}$. Thus in general, $L_i(y_1, y_2, \ldots, y_{i-1})f$ is defined if and only if $(y_1, y_2, \ldots, y_{i-1})$ is in a set $Q_i$.

We define the *radius of information* $R(N, A)$ in the same way as for a fully defined information operator:

We define the *radius of information* $R(N, A)$ in the same way as for a fully defined information operator:

$$R(N, A) = \sup_{(y_1,\ldots,y_{i-1}) \in Q_i \text{ all } i} R(S(N^{-1}(y) \cap A)).$$

**Lemma 3.8:** *For a fixed adaptive information operator $N$, a solution operator $S$, and a convex set $A$ in a separable Banach space $F$, the radius $R(N, SA)$ is supported on finite dimensional subsets. That is, for any convex $A$ there exist finite dimensional subsets $A_n \subset A$ such that $R(N, SA_n) \xrightarrow[n \to \infty]{} R(N, SA)$. The same remains true for partially defined information operators $N$.*

*Proof*: We give a proof of fully defined information operators $N$. Let $\{x_k\}_{k=1}^{\infty}$ be a dense set of points in $A$. Let $A_n$ be the affine subspace spanned by $\{x_k\}_{k=1}^{n}$ (i.e., the smallest affine subspace containing $\{x_k\}_{k=1}^{n}$), intersected with $A$. Let $A' = \bigcup_m A_m$. Note $A'$ is dense in $A$. We now show that $R(N, SA') = R(N, SA)$. Note that

$$R(N, SA') = \sup_{y \in \mathbb{R}^n} R(N, y, SA') \equiv \sup_{y \in \mathbb{R}^n} \text{Rad}(S(N^{-1})(y) \cap A'))$$

(3.5)

$$= \sup_{y \in \mathbb{R}^n} \text{Rad}(S(N^{-1}(y) \cap A)) = \sup_{y \in \mathbb{R}^n} R(N, y, SA) = R(N, SA),$$

where in the third equality we have used the fact that a set has the same radius as a dense subset, together with Lemma B above. We remark here that convexity of $A$ may be needed for the third equality to hold, since in fact this equality may be false for highly irregular sets $A$ whose intersections with the sets $N^{-1}(y)$ "miss" the dense subspace spanned by the $\{x_n\}$. Note that since $A'$ is dense in $A$, it follows that for any $y \in \mathbb{R}^d$, we have $N^{-1}(y) \cap A'$ is dense in $N^{-1}(y) \cap A$, by Lemma B. Further, since $A_n \underset{n \to \infty}{\uparrow} A'$, we have by Proposition 7, applied to sets of the form $N^{-1}(y) \cap A$, that

$$R(N, SA_n) = \sup_{y \in \mathbb{R}^d} \text{Rad}(S(N^{-1}(y) \cap A_n)) \underset{n \to \infty}{\uparrow} \sup_{y \in \mathbb{R}^d} \text{Rad}(S(N^{-1}(y) \cap A'))$$

$$= R(N, SA') = R(N, SA),$$

as desired; we have used Lemma 3.6 in evaluating the limit above. The proof in the case that N is partially defined is exactly the same as above. □

**Definition 3.3:** A *continuous* measure on a Banach space is one all of whose finite dimensional conditional and marginal measures have continuous density functions with respect to Lebesgue measure. A measure $\mu$ on a Banach space is *nonvanishing* if the conditional measure $\mu_y$ of $\mu$ restricted to any finite dimensional subspace satisfies $\lambda_y << \mu_y$, where $\lambda_y$ represents Lebesgue measure on that subspace and $<<$ denotes absolute continuity, and if the same holds for all finite dimensional marginal measures of $\mu$.

Examples of continuous measures on Banach spaces are Gaussian and elliptically contoured measures.

## 4. Proof of Theorems 2 and 3

**Lemma 4.1:** (*Lebesgue decomposition theorem for infinite dimensional spaces*): *Let $\mu$ be a Borel measure on a Banach space $F$. Then there exist unique nonnegative measures $\mu_1$ and $\mu_2$ such that $\mu = \mu_1 + \mu_2$, and such that for every finite dimensional subspace $G$, the conditional measure $\mu_1 G_a$ of $\mu_1$ on $G_a \equiv G + a$ (with respect to any complementary subspace $H$; here we assume $a \in H$) is absolutely continuous with respect to Lebesgue measure for a.e. $[\mu_{1H}]$ a, and $\mu_{2G}$ is singular with respect to Lebesgue measure, for a.e. $[g]$ a, where $\mu_{1H}$ is the marginal measure of $\mu_1$ on $H$. Further, all finite dimensional marginals of $\mu_1$ are absolutely continuous with respect to Lebesgue measure.*

More precisely, this Lemma states that given a pair of complementary subspaces $H$ and $G$ in $F$, with $\mu^H$ the marginal measure on $H$ and $\mu_G$ the conditional measure on $G$ with respect to the decomposition $F = G \oplus H$, then for almost every $G$ (with respect to the marginal measure $\mu^H$), $\mu_{1G}$ and $\mu_{2G}$ satisfy the properties mentioned above. Recall that since the conditional measures $\mu_{1G}$ and $\mu_{2G}$ are defined only on a set of $G$'s of full $\mu^H$-measure, this is the best possible statement of this type that can be made.

*Proof of Lemma 4.1*: Let $g$ be a nondegenerate Gaussian measure on $F$. Then by the Lebesgue decomposition theorem, the measure $\mu$ can be uniquely decomposed into a sum $\mu_1 + \mu_2$ in such a way that $\mu_1 << g$ and $\mu_2 \perp g$ (i.e., $\mu_2$ is singular with respect to $g$). It is easy to check that these two measures satisfy all the conditions of the Lemma. Namely, it is not difficult to show that the finite dimensional marginals of $\mu_1$ are a.c. with respect to those of $g$, and hence with respect to Lebesgue measure. To show the same of the finite dimensional conditional measures, note that by the above Lemma, since $\mu_1 \ll g$, it follows that the conditionals of $\mu_1$ are a.c. with respect to those of g, a.e. $[\mu_1]$. That the condtionals of $\mu_2$ are singular with respect to those of g and hence to Lebesgue measure follows immediately from the fact that $\mu_2$ is supported on a set $A \subset F$ of g-measure 0, and that therefore the cross sections $A_{x_1} = \{x = (x_1, x_2): x \in A\}$ have marginal g - measure 0 a.e. [g]. This proves that $\mu_1$ and $\mu_2$ have the properties desired. Unicity of $\mu_1$ and $\mu_2$ follows from unicity of their finite dimensional marginals, which are unique by the Lebesgue decomposition theorem. □

Recall that on a finite dimensional subspace, measures $\mu_n$ converge to a measure $\mu$ strongly if $\|\mu_n - \mu\|_1 \xrightarrow[n \to \infty]{} 0$, where the norm represents the sum of the absolute values of the total measures of the positive and negative parts of $\mu_n - \mu$.

**Def. 4.1:** Let $\mu$ be a Borel measure on a Banach space $F$. Then given a closed subspace $V \subset F$, we say that a subspace $T_1$ complementing $V$ is typical with respect to another subspace $T_2$ if the measure $\mu_{T_1}$, which is the conditional measure on $T_1$ with respect to $V$, has the property that the measure $\mu_{AT_1}(A \cdot)$ converges in $T_1 \oplus T_2$, and $\mu_{AT_1}$ denotes the conditional measure on $AT_1$ with respect to $V$.

Note that it is not difficult to show that if $T_1$ is typical, then this implies that if $P_A$ denote linear projections onto the subspaces $AT_1$ such that $\|P_A f - f\| \xrightarrow[A \to I]{} 0$ uniformly on bounded $f$-subsets of $T_1$, then $\mu_{AT_1} P_A \xrightarrow[A \to I]{} \mu_{T_1}$ strongly, where by definition $\mu_{AT_1} P_A(D))$ for $D$ is a measurable subset of $AT_1$.

**Lemma 4.2:** *If $\mu$ is a nonvanishing continuous measure on $\mathbb{R}^d$, then for any linear operator $N: \mathbb{R}^d \to \mathbb{R}^{d_1}$ and almost every $x_1 \in$ Range $N$, the subspace $N^{-1}(x_1)$ is typical. Thus*

$$\mu_{AN^{-1}(x_1)} P_A \xrightarrow[A \to I]{} \mu_{N^{-1}}(x_1)$$

*strongly, where $A$ represents a member of the (finite dimensional) space of affine transformations of $\mathbb{R}^d$, and $P_A$ represents the linear projection (relative to some independent subspace $G$) onto $AN^{-1}(x_1)$.*

*Proof*: Note that since $\mu$ can be represented by a density function $f$, we have for a $B$ measurable in $N^{-1}(x_1)$:

(15a)

$$\mu_{AN^{-1}} P_A(B) = c_A \int_{P_A(B)} f(x) d\lambda_{AN^{-1}(x_1)}(x) = c_A \int_B f(P_{A^x}) J_A d\lambda_{N^{-1}(x_1)}(x),$$

where $\lambda$ denotes conditional Lebesgue measure, $c_A$ is a constant which depends on the choice of the complementary $G$ and $A$ only, and $J_A$ denotes the (Lebesgue) Jacobian of the transformation $P_A$ from $N^{-1}(x_1)$ to $AN^{-1}(x_1)$. On the other hand, we have, writing $x = (x_1, x_2)$,

$$\int_G dx_1 \int_{N^{-1}(x_1)} dx_2 \, |f(P_A x) - f(x)| \xrightarrow[A \to I]{} 0,$$

since $|P_A x - x| \xrightarrow[A \to I]{} 0$, where above we have used standard facts about multtivariate functions (for example that the multivariate transformation group on $L^1$ generated by the set of affine transformations is continuous). Therefore, for almost every $x_1 \in G$, we have

$$\int_{N^{-1}(x_1)} dx \, |f(P_A x) - f(x)| \xrightarrow[A \to I]{} 0,$$

so that we conclude from (15a) that

$$\mu_{AN^{-1}(x_1)} P_A(B) = c_A \int_B f(P_{A^x}) J_A d\lambda_{N^{-1}(x_1)}(x)$$

$$\xrightarrow[A \to I]{} c_I \int_B f(x) J_I d\lambda_{N^{-1}(x_1)}(x)$$

$$= c_I \int_B f(x) d\lambda_{N^{-1}(x_1)}(x)$$

$$= \mu_{N^{-1}(x_1)}(B);$$

since this convergence can easily be shown to be uniform in the choice of the set $B$, it follows that we have the desired strong convergence. □

By convention, all $\mu$-measures on subspaces defined by inverse images of the operators $N_k$ will be assumed to be the appropriate conditional measures with respect to the marginal measures induced by the $N_k$. Recall we assume that all of our information operators consist of linear functionals of norm 1.

**Lemma 4.3:** *Suppose $\mu$ is a nonvanishing measure on a finite dimensional Banach space $F$, and $A$ is a convex set with $A_k \subseteq A$, and $\mu(A_k) \uparrow \mu(A)$. Let $N(y)(\cdot) = N_0(y)(\cdot)$ be a partially defined adaptive information operator of order $n$ on $F$ which is defined for a set $\mathcal{Q}$ of $y \in \mathbb{R}^n$ in such a way that $N$ is defined for a dense set of $f \in F$. Let $S$ be a linear operator from $F$ to a Banach space $G$. Let $\mu_y$ denote the conditional measure of $\mu$ restricted to $N^{-1}(y)$, with respect to the marginal measure defined by $N(y, \cdot)$. Let $N_k$ be adaptive information operators of order $n$ with the property that $N_k \xrightarrow[k \to \infty]{} N$ weakly, wherever $N$ is defined. Assume that for $k = 0, 1, 2, \ldots$, we have that for $y \in \mathcal{Q}$, $\mu(N_k^{-1}(y) \cap A_k)/\mu(N_k^{-1}(y) \cap A) \xrightarrow[k \to \infty]{} 1$ (where we assume $0/0 = 1$). Assume further that $N^{-1}(y)$ is typical with respect to $F$ for all $y \in \mathcal{Q}$. Then $\liminf_{k \to \infty} R(N_k, SA_k) \geq R(N < SA)$.*

Note that by weak convergence of the $N_k$ to $N$ we mean that for each component linear functional $L_{ik}(y)$ in $N_k$ we have $L_{ik}(y)(\,\cdot\,) \xrightarrow[k \to \infty]{} L_i(y)(\,\cdot\,)$ (generally in the weak topology of $F$) whenever $L_i(y)$ is defined.

*Proof of Lemma 4.3:* We will assume without loss that the measure $\mu$ is continuous (that is, that its finite dimensional conditionals are continuous). This can be done by Lemma 4.1, which shows that there is a nonvanishing continuous measure that is dominated by $\mu$. We will now adopt the notation $N_0 = N$.

Assume now that the conclusion of the lemma is false. Recall that the component functionals $L_{ik}(y_1, \ldots, y_{i-1}) \equiv L_{ik}(y)$ of the operators $N_k$ converge to $L_i(y_1, \ldots, y_{i-1})$ whenever the latter are defined (henceforth we omit the arguments $(y_1, \ldots, y_{i-1})$ in the linear functionals $L_i$ and $L_{ik}$. Since $F$ is finite dimensional, this convergence is uniform. Let $\rho > 0$, and $y \in \mathbb{R}^n$ be chosen such that

$$R(N, y, SA) > R(N, SA) - \rho, \tag{4.1}$$

and such that $y$ is in the range of the partially defined operator N, so that $N^{-1}(y)$ is typical. We claim that $R(N_k, y, SA_k) \xrightarrow[k \to \infty]{} R(N, y, SA)$. To see this, note first that we can take the set $N^{-1}(y)$ and consider projections (relative to a fixed complementary subspace $G$) onto $N^{-1}(y)$ of the sets $A_{ky} \equiv A_k \cap N_k^{-1}(y)$. We will assume that $G$ complements all of the subspaces $N_k^{-1}(y)$, which can be done since $N_k \xrightarrow[k \to \infty]{} N$.

Let $P_k$ denote the $G$-projection of $N_k^{-1}(y)$ onto $N^{-1}(y)$. Let $\mu_k$ denote the measure $\mu_{ky} P_k^{-1}$, where $\mu_{ky}$ denotes the conditional measure of $\mu$ on $N_k^{-1}(y)$ with respect to the complementary subspace $G$. Since $N_k^{-1}(y) \cap A$ converges in the maximal set distance sense to $N^{-1}(y) \cap A$, it follows that since the subspace $N^{-1}(y)$ is typical for $\mu$, that the measures $\mu_k$ converge in the sense of $L_1$ convergence of densities to $\mu_y$. Furthermore, if $A_{ky}^* = N_k^{-1}(y) \cap A$, and $A_k^* = P_k^{-1}(A_{ky}^*)$,

$$\mu_{ky}(A_{ky}^*)/\mu_y(A_k^*) \xrightarrow[k \to \infty]{} 1,$$

by the result of Lemma 4.2. On the other hand, by our assumptions,

$$\mu_{ky}(A_{ky})/\mu_{ky}(A_{ky}^*) \xrightarrow[k \to \infty]{} 1. \tag{4.2}$$

Note also that the measures $\mu_{ky} P_k$ converge strongly to $\mu_y$, by Lemma 4.2. Therefore, by (4.2) we have

$$\mu_y(P_k^{-1} A_{ky})/\mu_y(P_k^{-1} A_{ky}^*) \xrightarrow[k \to \infty]{} 1.$$

In addition, since $P_k^{-1} A_{ky} \subseteq P_k^{-1} A_{ky}^*$, we conclude that for any measurable set $C$, we also have

$$\mu_y(P_k^{-1}A_{ky} \cap C)/\mu_y(P_k^{-1}A_{ky}^* \cap C) \underset{k \to \infty}{\longrightarrow} 1,$$

so long as $\mu_y(P_k^{-1}A_{ky}^* \cap C) \geq \epsilon_1 > 1$ for some $\epsilon_1$ and sufficiently large $k$. But note also that

$$\mu_y(A_y)/\mu_y(P_k^{-1}A_{ky}^* \cap A_y) \underset{k \to \infty}{\longrightarrow} 1,$$

as can be seen from the fact that $A$ is a convex set, and that $|x - P_k x| \underset{k \to \infty}{\longrightarrow} 0$ for $x \in N^{-1}(y)$, together with the fact that $\mu$ is a continuous measure. Therefore, we conclude that

$$\mu_y(P_k^{-1}A_{ky} \cap A_y)/\mu_y(A_y) \underset{k \to \infty}{\longrightarrow} 1.$$

By Corollary 3.5, this implies that $\text{Rad}(SP_k^{-1}A_{ky}) \underset{k \to \infty}{\longrightarrow} \text{Rad}(SA_y)$, so that we have $\text{Rad}(SA_{ky}) \underset{k \to \infty}{\longrightarrow} \text{Rad}(SA_y)$. This proves that $R(N_k, y, SA_k) \underset{k \to \infty}{\longrightarrow} R(N, y, SA)$. Finally, the conclusion of the Lemma follows from (4.1) and the fact that $\rho > 0$ is arbitrary. $\square$

**Lemma 4.4:** *If $\mu$ is a nonvanishing probability measure on a convex set $A$ in a Banach space $F$, and if $A_k \subseteq A$, and $\mu(A_k) \uparrow \mu(A)$, then $R_n^{\text{wor}}(SA_k) \to R_n^{\text{wor}}(S, A)$, where $R_n^{\text{wor}}$ is either adaptive or nonadaptive Gelfand radius of order $n$, and $S: F \to G$ is any continuous solution operator, with $G$ a uniformly convex Banach space.*

*Proof:* Assume not. Then there exists a sequence of adaptive information operators $N_k$ such that

$$R(N_k, SA_k) \leq R_n^{\text{wor}}(SA) - \epsilon$$

for some fixed positive $\epsilon$. We assume as always that the components $L_i(y_{i-1}, y_{i-2}, \ldots, y_1)(x)$ have norm 1 as linear functionals. For convenience we extend the domain of $L_i$ to the set $\{y_{i-1}, \ldots, y_1; f \mid y_k \in \mathbb{R}; f \in F\} = \mathbb{R}^{i-1} \times F$; note that a priori the functionals $L_i$ are defined, for example, only for $y_1 \in \text{Ran}(L_1|_{F_0})$.

Define a set of linearly independent vectors $v_n$ which span $F$. Let $T_n = \text{span}\{v_1, \ldots, v_n\}$.

Let $L_{1k}$ be the first component of $N_k$. Clearly the sequence $\{L_{1k}\}_k \subseteq F^*$ has a weak-$*$ convergent subspace (note the sequence is also weakly convergent in $F^*$ since $F = F^{**}$ whenever $F$ is uniformly convex). This subsequence converges to some fixed functional $L_1 \equiv L_{1,0}$. Thus by taking subsequences, we may assume that $\{N_k\}$ has been chosen so that $L_{1k}$ converges in the weak-* topology. Now (after replacing $\{N_k\}$ by this subsequence) we will select a countable dense set of numbers $\mathbb{Q}_1$ as follows.

Define the measure $\mu_{L_1}$ to be the marginal measure $\mu L_1^{-1}$. We define our conditional measures below with respect to this marginal measure. Namely, define the conditional measures $\mu_{L_{1k,y}}$ to be the conditionals of the measure $\mu$ restricted to $L_{1k}^{-1}(y)$, with respect to the marginal measure $\mu_{L_1}$ (as opposed to $\mu_{L_{1k}}$).

Now define $Q_1 \subset \mathbb{R}$ to be the set of points such that for $y \in Q_1$, the ratio of conditional measures

$$W_k(y) \equiv \mu_{L_{1k},y}(L_{ik}^{-1}(y) \cap A_k)/\mu_{L_{1k},y}(L_{1k}^{-1}(y) \cap A) \xrightarrow[k \to \infty]{} 1. \tag{4.3}$$

(recall we have defined for convenience $0/0 = 1$). We will show (after possibly taking a subsequence in $k$) that we can conclude that $\mu_{L_1}(Q_1) = 1$.

Given $\eta > 0$ let $M_k(\eta)$ be the set of $y$ points such that $W_k(y) < 1 - \eta$. We claim that $\mu_{L_1}(M_k(\eta)) \xrightarrow[k \to \infty]{} 0$. For suppose not. Then for some $\eta > 0$, there is a $\rho > 0$ such that (after possibly taking subsequences in $k$) we have

$$\mu_{L_1}(M_k(\eta)) > \rho.$$

Consider

$$\mu(A_k) = \int_{\mathbb{R}^n} d\mu_{L_1}(y) \, \mu_{L_{1k},y}(A_k \cap L_{1k}^{-1}(y)),$$

where above all conditional measures $\mu_{L_{1k},y}$ are again understood to be chosen with respect to the marginal measure $d\mu_{L_1}$. By our assumption,

$$\mu(A_k) = \int_{\mathbb{R}^n} d\mu_{L_1}(y) \, W_k(y) \mu_{L_{1k},y}(A \cap L_{1k}^{-1}(y))$$

$$\leq \left(\int_{M_k(\epsilon)} + \int_{\sim M_k(\epsilon)}\right) d\mu_{L_1}(y) W_k(y) \mu_{L_{1k},y}(A \cap L_{1k}^{-1}(y))$$

$$\leq (1-\epsilon) \int_{M_k(\epsilon)} d\mu_{L_1}(y) \, \mu_{L_{1k},y}(A \cap L_{1k}^{-1}(y))$$

$$+ \int_{\sim M_k(\epsilon)} d\mu_{L_1}(y) \, \mu_{L_{1k},y}(A \cap L_{1k}^{-1}(y))$$

$$\leq (1-\epsilon)\mu_{L_1}(M_k(\epsilon)) + \int_{\sim M_k(\epsilon)} d\mu_{L_1}(y) \mu_{L_{1k},y}(A \cap L_{1k}^{-1}(y))$$

$$= (1-\epsilon)\mu_{L_1(M_k(\epsilon))+\mu_{L_1}(\sim M_k(\epsilon))}$$

$$= (1-\epsilon)\mu_{L_1}(M_k(\epsilon)) + 1 - \mu_{L_1}(M_k(\epsilon))$$

$$= -\epsilon \mu_{L_1}(M_k(\epsilon)) + 1$$

$$\leq 1 - \epsilon,$$

which contradicts the above assumption that $\mu(A_k) \xrightarrow[k \to \infty]{} \mu(A)$. Thus we have that for all $\epsilon > 0$, $\mu_{L_{1k}}(M_k(\epsilon)) \xrightarrow[k \to \infty]{} 0$, as desired. Thus the $W_k(y)$ converge to 1 in measure with respect to $\mu_{L_{1k}}$. By standard facts therefore, we can take a subsequence in $k$ which guarantees that $W_k(y)$ converges to 1 a.e. After taking this subsequence we have verified the above claim that $\mu_{L_1}(Q_1) = 1$.

Define $L_i \equiv L_{i0} \equiv \text{w-}\lim_{k \to \infty} L_i$. We will assume that all the finite dimensional subspaces $T_n$ as defined above are transverse to Ker $L_1$. If not, we can redefine the sequence $\{v_i\}$ so that no $v_i$ is contained in Ker $L_1$. For each subspace $T_n$ we permanently fix a complementary subspace $V_n$ which has sthe same transversality properties as assumed above for $T_n$. All conditional measures on the spaces $T_n$ will be assumed to be defined with respect to the complementary spaces $V_n$. Consider for fixex $T_m, T_{n'}$ the collection $\mathcal{A}$ of $a \in F$ with the property that $(T_n \cap \text{Ker } L_1) + a$ is typical with respect to $T_n$. Note that all conditional measures here are to be taken with respect to a fixed subspace $V_n \cap \text{Ker } L_1 \oplus \ell_1$ complementary to $T_n \cap \text{Ker } L_1$, where here $\ell$ is a one dimensional subspace complementary to $L_1$.

We claim that $\mathcal{A}$ has full measure in $F_0$. To see this, note that the set $T_n \cap \text{Ker } L_1 + a$ is finite dimensional, and if $a$ is momentarily restricted to be in a finite dimensional affine subspace $B + b \subset F$, with $B$ a subspace containing $T_n \oplus T_{n'}$ and $B + b$ endowed with a conditional measure relative to some fixed complementary subspace $B'$, then by Lemma 4.2, $T_n \cap \text{Ker } L_1 + a$ is typical with respect to $T_{n'}$ for almost all $a \in B$, when all measures are viewed as inherited from $B + b$. Note however that for almost all choices of $B$ (with respect to marginal measure $\mu_{B'}$ on $B'$) and almost all choices of $a$ within $B$, it is true that $T_n \cap \text{Ker } L_1 + a$ is typical with respect to $T_{n'}$ with conditional measures inherited from $B$ if and only if it is typical with respect to $T_{n'}$ with conditional measures inherited from $F_0$. Thus we conclude that for almost all $a \in F$, $T_n \cap \text{Ker } L_1 + a$ is typical with respect to $T_{n'}$ with respect to conditional measures inherited from $\mu$ on $F$, with respect to the complementary subspace $V_n \cap \text{Ker } L_1 \oplus \ell_1$ mentioned above.

It follows that for almost every $y$ with respect to the marginal measure $\mu_{L_1}$, it is true that for almost all $a \in L_1^{-1}(y)$, we have $T_n \cap \text{Ker } L_1 + a$ is typical with respect to $T_{n'}$, again viewing our conditional measures as inherited from $\mu$ on $F$. In addition, for almost all $a \in L_1^{-1}(y)$ the conditional measure in $T_n \cap \text{Ker } L_1 + a$ inherited from $F$ (using the complementary subspace $V_n \cap \text{Ker } L_1 \oplus \ell$) is proportional to the conditional measure in $T_n \cap \text{Ker } L_1 + a$ inherited from the conditional measure on $L^{-1}(y)$ (using the complementary subspace $V_n \cap \text{Ker } L_1$). Therefore, we conclude that for almost every $a \in L^{-1}(y)$ (with respect to the conditional measure $\mu_{L_1,y}$), $T_n \cap \text{Ker } L_1 + a$ is typical with respect to $T_{n'}$, using conditional measures inherited from the conditional measure $\mu_{L_1,y}$, relative to the complementary subspace $V_n \cap \text{Ker } L_1$.

Let $\mathcal{A}_y(n, n')$ be the set of $a \in L_1^{-1}(y)$ such that $T_n \cap \text{Ker } L_1 + a$ is typical with respect to $T_{n'}$. We have shown above that $\mu_{L_1,y}(L_1^{-1}(y) \sim \mathcal{A}_y(n, n')) = 0$. Let

$$\mathcal{A}_y = \bigcap_{n,n'} \mathcal{A}_y(n, n').$$

Then a standard countability argument shows that

$$\mu_{L_1,y}(L_1^{-1}(y) \sim \mathcal{A}_y) = 0,$$

i.e., that for almost every $y$ and almost every $a \in L_1^{-1}(y)$, the set $L^{-1}(y) \cap T_n + a$ is typical with respect to $T_{n'}$ for all $n, n'$.

Now choose the set $\mathbb{Q}_1 \subseteq Q'_1$ to be a countable dense set in $L_1(F)$, which is possible since $Q'_1$ has full measure with respect to the nonvanishing measure $\mu_{L_1}$ on $L_1(F)$.

For each $y_1 \in \mathbb{Q}_1$, we may assume that $L_{2k}(y_1)$ converges weakly to $L_2(y_1) \equiv L_{2,0}(y_1)$ (by a Cantor diagonal argument for subsequences).

We now restrict our attention to $L_1^{-1}(y_1)$ for some fixed $y_1$. Note that the measure on this set is the conditional $\mu_{L_1,y}$; the new linear functional $L_2$ defined on $L_1^{-1}(y)$ defines a new marginal measure *with respect* to the conditional $\mu_{L_1,y}$. We will denote this marginal measure with respect to $L_2$ within $L_1^{-1}(y)$ by $\mu_{L_1,y;L_2}$. and its precise definition is

$$\mu_{L_1,y;L_2} \equiv \mu_{L_1;y} L_2(y_1)^{-1}, \tag{4.4}$$

where $L_2(y_1)(\,\cdot\,)$ in (4.4) is by assumption restricted so that its domain $\cdot$ consists only of elements of $L_1^{-1}(y_1)$.

Furthermore, we can show using the exact same arguments as before that for any given $y_1 \in \mathbb{Q}_1$, there exists a set $Q_{2,y_1} \subset \mathbb{R}$ of full marginal measure with respect to $\mu_{L_1,y;L_2}$ such that for $y_2 \in Q_{2,y_1}$,

$$\mu_{y_1,y_2}(L_{1,2}^{-1}(y_1,y_2) \cap A_k)/\mu_{y_1 y_2}(L_{1,2}^{-1}(y_1,y_2) \cap A) \xrightarrow[k \to \infty]{} 1, \tag{4.5}$$

where we have adopted the notation $L_{i_1\ldots i_j} \equiv (L_{i_1}, L_{i_2}, \ldots, L_{i_j})$. Here, the measure $\mu_{y_1 y_2}$ denotes the conditional measure on $L_{1,2}^{-1}(y_1, y_2)$ with respect to the marginal defined by $\mu_{L_{1,2}} = \mu_{L_{1,2}}^{-1}$. To see why (4.5) holds, note that the conditional measure $\mu_{L_1,y_1}$ is continuous, and so by the same arguments as those which established (4.3), equation (4.5) follows.

We can again further choose $Q_{1,y_1}$ so that for almost every $y_2$ (with respect to the marginal measure $\mu_{L_1,y;L_2}$), the set $\mathcal{A}_{y_1 y_2}$ has full measure in $L_{1,2}^{-1}(y_1, y_2)$ (with respect to the conditional measure $\mu_{y_1 y_2}$). Here $\mathcal{A}_{y_1 y_2}$ is defined as the set of all $a \in L_{1,2}^{-1}(y_1, y_2)$ such that $(\mathrm{Ker}\, L_{1,2} \cap T_n) + a$ is typical with respect to $T_{n'}$ for all $n, n'$, with respect to conditional measures inherited from the conditional measure $\mu_{y_1 y_2}$.

For notational convenience, we assume that the sets $Q_{2,y_1}$ are defined in such a way that whenever $L_{1,2}^{-1}(y_1, y_2)$ is empty, $y_2$ is automatically included in $Q_{2,y_1}$.

Now let

$$Q_2 = \bigcap_{y_1 \in \mathbb{Q}_1} Q_{2,y_1},$$

so that $Q_2$ is of full measure, and let $\mathbb{Q}_2 \subset Q_2$ be countable set which is dense in $Q_2$. We continue in this way until we have a collection $\mathbb{Q}_1, \mathbb{Q}_2, \ldots, \mathbb{Q}_n$ of countable dense sets in $\mathbb{R}$ such that for any $y \in \mathcal{Q} = \mathbb{Q}_1 \times \ldots \times \mathbb{Q}_n$, we have that $L_i(y_1, \ldots, y_{i-1})$ converges weakly in $i$, and such that $\mu(N^{-1}(y) \cap A_k)/\mu(N^{-1}(y) \cap A) \xrightarrow[k \to \infty]{} 1$, and also that for any $y \in \mathcal{Q}$ and any $n, n'$, $(\mathrm{Ker}\, N(y) \cap T_n) + a$ is typical with respect to any $T_{n'}$, for all $a \in N^{-1}(y)$ in a set of full conditional measure, where $y = (y_1, \ldots, y_n)$.

Consider now the partially defined adaptive information operator $N(f)$ with component linear functionals $L_i(y_1, y_2, \ldots, y_{i-1}; f)$, as described above. Since $L_i(y)$ is defined only for $y_j \in \mathbb{Q}_j$ for $j < i$, it follows that $N$ is only a partially defined information operator. By Lemma C, we note that the radius of information with respect to $N$ is approximately supported on a finite dimensional subset, and furthermore, from the proof of the Lemma, it can be seen that such a set can be chosen to be of the form $T_n$, since $\bigcup_n T_n$ is dense in $F$ and hence in any fibre $N^{-1}(y)$, $y \in \mathcal{Q}$. More precisely, given $a > 0$, there exists a finite dimensional subspace $T_n \subseteq F$ such that if $A_{T_n} = A \cap T_n$ then $R(N, SA_{T_n}) > R(N, SA) - a$.

We will select the subspace $T_n$ more precisely as follows. First note that $R(N, SA_T)$ is a continuous function of the subspace $T$ in that $R(N, SA_{T+a})$ is a continuous function of $a \in F$ in the interior of its support, as follows from an argument like that for Proposition 5 above (note that here T is a finite dimensional set, though the proof of the result is exactly the same in this case). Note in particular that by construction, the subspaces $T_n$ are transverse to $N^{-1}(y)$, which allows us to use an argument like that in Proposition 5.

We will choose $T$ as follows. First, choose a finite dimensional subspace $T_n$ from the above collection for which $R(N, SA_{T_n}) > R(N, SA_T) - \alpha$ for some pre-determined $\alpha > 0$. Now consider the collection of spaces of the form $T_a = T_n + a$, for $a \in F$. Let

$$D_0 = \{a \in F \mid \mu_a(A_{kT_a}) \xrightarrow[k \to \infty]{} \mu_a(A_{T_a})) \text{ and } T_a \text{ is typical with respect to all } T_m\}, \quad (4.6)$$

where $\mu_a$ denotes the conditional measure on $T_a$ with respect to the complementary subspace $V_n$. Then note that $D_0$ has full measure with respect to $\mu$, by the same arguments as used previously to show that the sets $Q_i$ have full measure.

Recall we have constructed the subspaces $T_n$ (with readjustments of the selection of the $v_n$ if necessary) so that the $T_n$ are all transverse to the subspaces $N^{-1}(y)$ for $y \in \mathcal{Q}$. This has been possible because the set of all possible $y \in \mathcal{Q}$ and hence the collection of subspaces $N^{-1}(y)$ is countable.

Now let $\{y^i\}$ denote an enumeration of $\mathcal{Q}$, and define $D_i$ by

(4.7)
$$D_i = \{a \in F \mid \mu_A(N^{-1}(y^i) \cap A_{kT_a}) \xrightarrow[k \to \infty]{} \mu_A(N^{-1}(y^i) \cap A_{T_a}) \text{ and } T_a \text{ is typical for all } T_m\}$$

Again since all of the $D_i$ have full measure, we let $D = \bigcap_i D_i$, and $D$ also has full measure. Thus we can shoose an $a$ arbitrarily close to any given number, $a_1$ in such a way that $T_n$ satisfies the condition inside of (4.7) above, for all $y \in \mathcal{Q}$. Thus by the continuity of the radius of information, we choose $a$ from the dense set of values such that the condition in (4.7) is satisfied, and also so that

$$R(N, SA_{T_a}) > R(N, SA) - a. \quad (4.8)$$

Define $T = T_a$ for this choice of $a$.

We now restrict to the finite dimensional affine subspace $T$. Since $\alpha$ in (4.8) is arbitrary, it suffices to show that

$$\liminf_{k \to \infty} R(N_k, SA_k) \geq \liminf_{k \to \infty} R(N_k, SA_{kT}) \geq R(N, SA_T), \quad (4.9)$$

so that our problem is now reduced to a finite dimensional one in the affine subspace $T$. Note now that for each $y \in \mathcal{Q}$, $\mu(N^{-1}(y) \cap A_{kT})/\mu(N^{-1}(y) \cap A_T) \xrightarrow[k \to \infty]{} 1$, and also that $\mu(A_{kT})/\mu(A_T) \xrightarrow[k \to \infty]{} 1$, so that by Lemma 4.3 for finite dimensional spaces, we conclude that indeed (4.9) holds, as desired. Letting $\alpha \to 0$ we concliude that

$$\liminf_{k \to \infty} R(N_k, SA_k) \geq R(N, SA) \geq R_n^{\text{wor}}(SA).$$

By the continuity of the local radius of information it is easy to show that even though $N$ is partially defined, the fact that it is defined for a dense set of $y$ means that it has an everywhere defined extension $\tilde{N}$ with the same radius of information, so that

$$R_n^{\text{wor}}(SA) \leq R(\tilde{N}, SA) = R(N, SA) \leq \liminf_{k \to \infty} R(N_k, SA_k).$$

Therefore,

$$\liminf_{k \to \infty} R(N_k, SA_k) \geq \text{Rad}_n^{\text{wor}}(SA),$$

giving us the desired contradiction of the assumption (34) at the beginning of the proof. □

*Proof of Theorem 3:* Assume that the result is false. Then we have

$$\text{Rad}_\delta^{\text{prob}}(n) \xrightarrow[\delta \to 0]{} R_0 < \text{Rad}^{\text{wor}}(n).$$

Thus there exists a sequence $\delta_k \xrightarrow[d \to 0]{} 0$ such that

$$\text{Rad}_{\delta_k}^{\text{prob}}(n) < \text{Rad}^{\text{wor}}(n) - \gamma$$

for some $\gamma > 0$. But note that

$$\text{Rad}_{\delta_k}^{\text{prob}}(n) = \inf_{A_k \subset A; \mu(A_k) < 1-\delta} \text{Rad}_n^{\text{wor}}(SA_k).$$

Therefore, there exists a sequence of sets $A_k \subset A$ such that $\mu(A_k) \geq 1 - \delta_k$, and

$$R_n^{\text{wor}}(SA_k) \leq R_n^{\text{wor}}(SA) - \gamma/2.$$

However, note that $R_n^{\text{wor}}(SA)$ is just the adaptive Gelfand $n$-radius of $SA$, and thus this statement contradicts Lemma 4.4. Therefore, we conclude that the statement of the theorem holds for the $\delta \to 0$ convergence of probabilistic radii of information $R_\delta^{\text{prob}}(n)$ to the worst case radius $R^{\text{wor}}(n)$.

We now need to translate these results into ones on cardinality of information. Recall that the cardinality of our problem $S$ is defined by

$$\text{card}^\delta(S, \epsilon) \equiv C^\delta(\epsilon) = \inf\{n : R_\delta^{\text{prob}}(n) < \epsilon\},$$

Now we wish to show that for any $\epsilon$,

$$C^\delta(\epsilon) \xrightarrow[\delta \to 0]{} C(\epsilon) \equiv \text{worst case complexity}.$$

Now note that

$$C^\delta(\epsilon) = \inf\{n : R_\delta^{\text{prob}}(n) \leq \epsilon\}.$$

Since the functions $R_n^\delta$ are monotonically decreasing in $n$ and increasing in $\delta$, it follows without too much difficulty that according to our definitions, $C^\delta(\epsilon) \xrightarrow[\epsilon \to 0]{} C(\epsilon)$, as desired, at any continuity point $\epsilon$ of the function $C(\epsilon)$. □

*Proof of Theorem 2:* The proof of Theorem 2 follows in the same way as the proof of Lemma 3.4 above and the proof of Theorem 3. The only difference is that the sequence $N_k$ of operators in the proof of Lemma 3.4 is replaced by the single operator $N$. □

*Proof of Theorem 1:* This theorem is now a corollary to Theorems 2 and 3. □

# 5. The $p \to \infty$ Limit For $p$-average Case Complexity.

In this section we will prove the statememts of Theorem 10. For convenience here we will define the $L^p$-*expectation* of a random variable $X$ by

$$E_p(X) = E(|X|^p)^{1/p},$$

where $E$ denotes ordinary mathematical expectation. The average case complexity and radius of information are generally defined in terms of mean square error, i.e., $E_2$. The local radius, for example, can be written in the form (see [TWW, §6.2])

$$R_{\text{avg}}(N, y) = \inf_{h \in G} E_2(\|h - Sf\| \,|\, y) \equiv \inf_{h \in G} \left( \int_{N^{-1}(y)} \|h - Sf\|^2 \mu(df|y) \right)^{1/2} \quad (5.1)$$

where $\mu(df|y)$ is the conditional measure of $\mu$ on the set $N^{-1}(y)$, with respect to the marginal measure $\mu N^{-1}$ on $Y = \mathbb{R}^n$. To define the average radius of information for $N$, we have

$$R_{\text{avg}}(N) = E_2(R_{\text{avg}}(N, y)) = \left( \int R_{\text{avg}}(N, y)^2 \, d\mu_1(y) \right)^{1/2}, \quad (5.2)$$

where $\mu_1 = \mu N^{-1}$ is the image of the measure $\mu$ under $N$. When square norms are replaced by $L^p$ norms, we have what is called the $L^p$ average local radius of information, or $R_p^{\text{avg}}(N, y) = \inf_{h \in G} E_p(\|h - Sf\| \,|\, y)$, with the radius of information $N$ defined by

$$R_p^{\text{avg}}(N) = E_p(R_p^{\text{avg}}(N, y))$$

where the expectation $E_p$ above is taken with respect to the measure $\mu_1 = \mu N^{-1}$. Thus the $p$-average radius can also be written as an infimum over functions $h$ given by

$$R_p^{\mathrm{avg}}(N) = \inf_{h:\mathbb{R}^n \to G} \left( \int_F \|Sf - h(Nf)\|^p \, d\mu(f) \right)^{1/p}.$$

Note that measurability issues regarding the function $h$ do not arise since it can be shown (see [TWW], §6.2) that the local radius of information in the $L^p$ setting (defined analogously to (5.1) with 2- norms replaced by $p$-norms) for $p < \infty$ is a measurable function. The set of $h$ over which the infimum is taken therefore can simply be chosen to be simply the set of $h$ for which the locally defined function

$$R_p^{\mathrm{avg}}(N, y) = \inf_{h \in G} E_p(\|h - Sf\| \,|\, y)$$

is measurable in $y$ (note that an equation analogous to (5.2) can also be used to define the $p$-average radius of information).

Finally, the overall $p$-average radius of information is given by

$$R_p^{\mathrm{avg}} = \inf_{N \in \mathcal{N}} R_p^{\mathrm{avg}}(N),$$

where $\mathcal{N}$ denotes the class of admissible information operators $N : F \to Y = \mathbb{R}^n$.

In order to study the limit as $p \to \infty$ of $R_{\mathrm{avg}}(N)$, we require the estimate

$$R_p^{\mathrm{avg}}(N) = \inf_{h:\mathbb{R}^n \to G} \left( \int_F \|Sf - h(Nf)\|^p \, d\mu(f) \right)^{1/p}$$

$$\geq \inf_{h:\mathbb{R}^n \to G} \left( \left( \int_{\|Sf - h(Nf)\| \geq R_\delta^{\mathrm{prob}}(N)} \|Sf - h(f)\|^p \, d\mu(f) \right)^{1/p} \right)$$

$$\geq \inf_{h:\mathbb{R}^n \to G} \left( \left( R_\delta^{\mathrm{prob}}(N)^p \, P(\|Sf - h(Nf)\| \geq R_\delta^{\mathrm{prob}}(N)) \right)^{1/p} \right)$$

$$= \inf_{h:\mathbb{R}^n \to G} R_\delta^{\mathrm{prob}}(N) \left( P\left(\|Sf - h(Nf)\| \geq R_\delta^{\mathrm{prob}}(N)\right) \right)^{1/p}$$

$$\geq R_\delta^{\mathrm{prob}}(N) \, \delta^{1/p},$$

where $R_\delta^{\mathrm{prob}}(N)$ is the probabilistic radius of information for probability $1 - \delta$. Thus taking infima over operators $N$ of cardinality $n$, we have

$$R_p^{\mathrm{avg}}(n) = \inf_N R_p^{\mathrm{avg}}(N) \geq \inf_N R_\delta^{\mathrm{prob}}(N) \, \delta^{1/p} = R_\delta^{\mathrm{prob}}(n) \, \delta^{1/p}.$$

Note that we are defining the $p$-average radius of order $n$ as an infimum over information operators $N$ of fixed cardinality $n$; see the remark after the statement of Theorem 11.

Further, letting $p \to \infty$,

$$\lim_{p \to \infty} R_p^{\text{avg}}(n) \geq R_\delta^{\text{prob}}(n) \tag{5.3}$$

for all $\delta$. Letting $\delta \to \infty$, we conclude that

$$\lim_{p \to \infty} R_p^{\text{avg}}(n) \geq R^{\text{wor}}(n), \tag{5.4}$$

since we have showed that the $\delta \to 0$ limit of the right side of (5.3) is the right side of (5.4).

On the other hand, noting that for any random variable $X$, we have

$$E(X^p)^{1/p} \leq \|X\|_\infty$$

we conclude that for each $y$,

$$R_p^{\text{avg}}(N, y) = \inf_{h \in G} E_p(\|h - Sf\| \,|\, y)$$

$$= \inf_{h \in G} \left( \int_{N^{-1}(y)} \|h - Sf\|^p \mu(df \,|\, y) \right)^{1/p}$$

$$\leq \inf_{h \in G} \sup_{f \in N^{-1}(y)} \|h - Sf\|$$

$$= R^{\text{wor}}(N, y)$$

Further,

$$R_p^{\text{avg}}(N) = E_p(R_p^{\text{avg}}(N, y)) \leq E_p(R^{\text{wor}}(N, y)) \leq \|\mathbf{R}^{\text{wor}}(N, y)\|_{\infty, y} \leq R^{\text{wor}}(N),$$

and taking suprema over (adaptive or nonadaptive) information operators $N$, we conclude

(7.11) $$R_p^{\text{avg}}(n) \leq R_n^{\text{avg}}(n). \tag{5.5}$$

Combining (5.4) and (5.5), we have $\lim_{n \to \infty} R_p^{\text{avg}}(n) = R_n^{\text{wor}}(n)$, as desired. The further identities stated in Theorem 11 all follow from the above identities as well.

## 6. Model of Complexity and the Proof of Theorem 13

Our model of complexity here assumes that an algorithm $\phi$ for solving the problem $S$ may have a complexity which depends arbitrarily on the solution operator $N$ and the information $y$. Before considering the proof of Theorem 12, we will require a lemma regarding the relationship of Gelfand $n$-widths and compactness (the converse of this lemma appears in [P]).

**Lemma 6.1:** *Let $G$ be a Banach space, and let $A \subset G$ be convex and balanced. Then if the Gelfand $n$-widths $D^k(A)$ of $A$ converge to $0$, $A$ is compact.*

*Proof:* It is easy to check that if $A$ is convex and balanced, then our hypothesis implies that there exists a sequence of linear functionals $L_k$ of norm 1 such that if $N_k = (L_1, L_2, \ldots, L_k)$, then $R(N_k, A) \leq D^k(A) \xrightarrow[k \to \infty]{} 0$. Let $\{a_n\}$ be a sequence in $A$. Then using a diagonalization argument, there is a subsequence $a_{n_l}$ such that for all $j$, $L_j(a_{n_l})$ converges monotonically to a constant $\ell_j$ as $l \to \infty$, with $|L_j(a_{n_l}) - \ell_j| < c_{lj}$, and $c_{l_j} \xrightarrow[l \to \infty]{} 0$ for each $j$. Without loss we may assume that $c_{l_j}$ decrease monotonically in $l$ and in $j$. Note that for each finite $j$ and $\epsilon > 0$ there is a $\delta_j(\epsilon)$ such that if $|L_k(a)| \leq \delta_j(\epsilon)$ $(1 \leq k \leq j)$, then $d(a, \text{Ker } N_j) < \epsilon$, where $d$ denotes distance in $F$. We will assume without loss that $\delta_j(\epsilon)$ is a decreasing function of $j$. Choose $c_{l_j}$ so that $c_{l_j} \leq \frac{1}{2} \delta_l(1/l)$ for $l \geq j$. This implies that for $l_1, l_2 \geq j$,

$$|L_k(a_{nl_1}) - L_k(a_{n_{l_2}})| \leq 2 \, c_{jk} \leq \delta_j(1/j) \quad (k = 1, \ldots, j),$$

so that $d\,((a_{n_{l_1}} - a_{n_{l_2}}), \text{Ker } N_j) \leq 1/j$. Therefore, for $l_1, l_2 \geq j$,

$$|a_{n_{l_1}} - a_{n_{l_2}}| \leq \text{diam (Ker } N_j) + 1/j \xrightarrow[j \to \infty]{} 0,$$

so that the sequence is indeed Cauchy. Thus $A$ is compact, as desired. $\square$

*Proof of Theorem 13*: Note that since the $\epsilon$-complexity is assumed to be defined for all $\epsilon > 0$ (however small), it follows that the adaptive radius of information $R^{\text{wor}}(n) \xrightarrow[n \to \infty]{} 0$, and hence that the same is true of the nonadaptive radiius and thus diameter of information. Therefore, the Gelfand $n$-widths of the set $SF_0$ converge to 0, so that $SF_0$ is is compact. Therefore, weak convergence properties in $F$ are translated under $S$ into strong convergence properties. Note in particular that if a sequence of information operators $N_n$ converges weakly to $N$ in $F$, then $E_n = S(\text{Ker } N_n \cap F_0)$ converges partially in the sense of maximal set distance to $E = S(\text{Ker } N \cap F_0)$. Specifically, the distance

$$\sup_{g \in E} d(g, E_n) \xrightarrow[n \to \infty]{} 0 \qquad (6.1)$$

This can be seen by the fact that we can restrict attention to a finite dimensional subspace $V$, and certainly (6.1) holds on restriction to $V$ in both $E$ and $E_n$. Choosing $V$ as increasingly better approximations to $SN^{-1}(y) \cap F_0$ gives (6.1). We will use the compactness of $SF_0$ and thus finite dimensional approximability here in this way.

For each $\delta > 0$, and for each $c$, define $(N_\delta, \phi_\delta)$ as an information operation and algorithm such that

$$e_\delta(N_\delta, \phi_\delta) \leq \epsilon,$$

for a fixed $\epsilon > 0$, and

$$C(N_\delta, \phi_\delta) = \text{comp}_\delta(\epsilon) + \delta,$$

where $C(N, \phi)$ denotes the total complexity $\text{comp}(N, \phi)$ (equation (1.6)). Now without loss of generality assume that we take a further subsequence above so that the operators $N_{\delta_k}(y, \cdot)$ all have the same cardinality $n$ and converge weakly to an operator $N$ for $y$ in a dense subset $\mathcal{Q} \in \mathbb{R}^d$, constructed as in the proof of Lemma 4.4 in such a way that all of the conditions in that proof are satisfied.

Define the local worst case error by::

$$e(N, \phi, y) = \sup_{Nf=y} \|\phi(y) - Sf\|.$$

We have

$$\mu\{f \in F_0 : e(N_{\delta_k}, \phi_{\delta_k}, N_{\delta_k}f) \leq e\} \geq 1 - \delta_k \xrightarrow[k \to \infty]{} 1.$$

Therefore, as in the proof of Lemma , we can choose our set $\mathcal{Q}$ so that for $y \in \mathcal{Q}$,

$$\mu_{ky}\{f \in F_{ky} : e(N_{\delta_k}, \phi_{\delta_k}, y) \leq \epsilon\} \xrightarrow[k \to \infty]{} 1, \qquad (6.2)$$

where $\mu_{ky}$ is the conditional measure on $N_{\delta_k}^{-1}(y)$ with respect to the marginal $\mu N^{-1}$, and $F_{ky} \equiv F_0 \cap N_{\delta_k}^{-1}(y)$.

For a fixed $y \in \mathcal{Q}$, consider the local worst case error $e(N, \phi_{\eta_k \delta_k}, y)$. We claim that

$$\liminf_{n \to \infty} e(N, \phi_{\delta_k}, y) \leq \epsilon. \qquad (6.3)$$

For suppose not. Then we can assume (possibly by taking subsequences) that

$$e(N, \phi_{\delta_k}, y) > \gamma > \epsilon, \qquad (6.4)$$

for all $k$. Thus for each $k$ there exists $f_k \in N^{-1}y$ such that

$$\|\phi_{\delta_k} N(y) - Sf_k\| > \gamma,$$

However, as in the proof of Lemma 4.4, we can select a finite dimensional subspace $V$ (denoted by $T$ in that Lemma) on which the radius of $S(N^{-1}(y) \cap F_0)$ is essentially supported, so that

$$R(N, y, SF_{0V}) > R(N, y, F_0) - \alpha$$

for some small $\alpha > 0$, where henceforth the subscript $V$ on a set denotes that set intersected with $V$. We can, further, assume without loss that $F$ is transverse to $N^{-1}(y)$ and $N_{\delta_k}^{-1}(y)$ for all $k$ and $y \in \mathcal{Q}$, as in the proof of Lemma 4.4. According to our choice of $y$, we have because of our restriction to finite dimension (see the proof of Lemmas 4.3, 4.4)

$$R(N_{\delta_k}, y, SF_{0V}) \xrightarrow[k \to \infty]{} R(N, y, SF_{0V}).$$

Further, since $SF_0$ is pre-compact, we can choose the finite dimensional set $V$ so that the maximal set distance

$$\delta(SF_{0Vy}, SF_{0y}) < \alpha.$$

It is then not difficult to show (using the measure theoretic arguments in the proof of Lemma 4.4) that

$$\delta(SF_{0y}, S(N^{-1}_{\eta_k \delta_k}(y) \cap A_k)) \xrightarrow[k \to \infty]{} 0.$$

Therefore, the worst case error

$$e(N, \phi_{\delta_k}, y) = \sup_{f \in N^{-1}} \|\phi_{\delta_k}(y) - Sf\|$$

$$\leq \sup_{f \in N^{-1}(y) \cap V} \|\phi_{\delta_k}(y) - Sf\| + \alpha \qquad (6.5)$$

$$\equiv e_V(N, \delta_k, y).$$

On the other hand, we can select $V$ so that $N^{-1}(y) \cap V$ is typical for $\mu$ with respect to $V$. Then arguments like those in the proof of Lemma 4.4 show that (since the measure $\mu$ may again be assumed without loss of generality to have continuous restrictions to finite dimensional subspaces)

$$\mu_{kyV} P_{kV} \xrightarrow[k \to \infty]{} \mu_{yV}, \qquad (6.6)$$

where $\mu_{kyV}$ denotes the conditional of $\mu$ on $N^{-1}_{\delta_k}(y) \cap V$, and $P_{kV}$ denotes the projection of $N^{-1}(y) \cap V$ onto $N^{-1}_{\delta_k}(y) \cap V$, and $\mu_{yV}$ is the conditional on $N^{-1}(y) \cap V$, the above projections and conditionals taken with respect to a fixed complementarity subspace of $N^{-1}(y) \cap V$. The convergence above is again in the strong topology in the space of signed measures. Note now that by (6.2), we can select $V$ so that in addition to the properties above, it holds that

$$\mu_{kyV}\{f \in F_{kyV} : e_V(N_{\delta_k}, \phi_{\delta_k}, y) \leq \epsilon\} \xrightarrow[k \to \infty]{} 1, \qquad (6.7)$$

where $\epsilon_V$ is as in (6.5). On the other hand, by (6.6), it follows from (6.7) that

$$\mu_{yV}\{f \in F_{yV} : \epsilon_V(N, \phi_{\delta_k}, y) \leq \epsilon\} \xrightarrow[k \to \infty]{} 1$$

(note that $N_{\delta_k}$ can now be replaced by $N$). Since the measure $\mu_{yV}$ is continuous and we are in a finite dimensional setting, it follows that given any $\beta > 0$, it is true that for all $y \in \mathcal{Q}$ and $k$ sufficiently large,

$$\epsilon_V(N, \phi_{\delta_k}, y) < \epsilon + \beta.$$

It follows from the above that letting $\alpha$ and $\beta \to 0$, we can conclude that for any $y \in \mathcal{Q}$ and $\eta$, if $k$ is sufficiently large,

$$e(N, \phi_{\delta_k}, y) < \epsilon + \eta, \tag{6.8}$$

since $V$ can be chosen to approximate $S(N^{-1}(y) \cap F_0)$ arbitrarily well in the sense of maximal set distance. This contradicts (6.4), proving (6.3).

Recall that the set $\mathcal{Q}$ was selected in such a way that

$$\mathcal{Q} = \{y \in \mathbb{R}^n : y_1 \in \mathbb{Q}_1, \, y_2 \in \mathbb{Q}_2(y_1), \, y_3 \in \mathbb{Q}_3(y_1, y_2), \ldots, y_n \in \mathbb{Q}_n(y_1, y_2, \ldots, y_{n-1})\},$$

where $\mathbb{Q}_k(y_1, \ldots, y_{k-1})$ is a countable dense subset of the real numbers. Let us now consider the local error as a function of $y$. Let $y = (y_1, y_2, \ldots, y_n)$, and fix $y^{(n-1)} = (y_1, \ldots, y_{n-1})$. For $k \leq n$, define

$$\mathcal{Q}_k = \{(y_1, y_2, \ldots, y_k) : y_i \in \mathbb{Q}_i(y_1, \ldots y_{i-1})\}.$$

For $y^{(k)} \in \mathcal{Q}_k$, let

$$\mathcal{Q}_{y^{(k)}} = \{y \in \mathbb{R}^{n-k} : (y^{(k)}, y) \in \mathcal{Q}\}.$$

We then know that for $y_n \in \mathcal{Q}_{y^{(n-1)}}$,

$$e(N, \dot{\phi}_{\delta_k}, (y^{(n-1)}, y_n)) < \epsilon + \gamma \tag{6.9}$$

if $k$ is sufficiently large. Here and henceforth, the dot above the operator $\phi(\,\cdot\,)$ generically indicates that $\phi(\,\cdot\,)$ has been replaced by $\phi(a\,\cdot\,)$, where $a < 1$ is sufficiently close to 1 so that (8.18)(6.9). (or any other identity under consideration) is still valid (note that our assumption is that the $\phi_{\delta_k}$ are continuous, uniformly in $k$), but $a$ is sufficiently small that $\dot{\phi}_{\delta_k}$ has the property that for all $f$ in $F_0$, $N(f)$ is in the domain of $\dot{\phi}_{\delta_k}$. The accomplishment of this may require that $k$ be increased, which does not create any difficulty, since we assume $k$ is so large that oour identities hold not only for $k$ individually, but for the given $k$ and all $k$ greater than it. Note that this involves no change in complexity, since the multiplier $a$ in the argument of $\phi$ may be interpreted as a multiplier of the adaptive information operator $N$, and this does not change the cardinality of the operator $N$, the scaling of whose components is initially arbitrary in any case.

Define $T(y^{(n-1)}) = \{f \in F_0 : Nf = (y^{(n-1)}, y_n) \text{ for some } y_n\}$. Then $N$ is purely linear in $T(y^{(n-1)})$. Note that therefore if $N$ is only partially defined in $y^{(n-1)}$, there is a unique linear extension of $N$ in $T(y^{(n-1)})$ so that $N$ is linear there. Thus we may assume that for each $y^{(n-1)}$, $N$ is defined and linear in all of $T(y^{(n-1)})$, as opposed to just $\mathcal{Q} \cap y^{(n-1)})$. We know that for any $y \in \mathcal{Q}$, (8.17)(6.8) holds for $k$ sufficiently large, so by our allowed assumption of a uniformly bounded modulus of continuity for the $\phi_{\delta_k}$, it follows that for $k$ sufficiently large it is true that for *all* of $f \in T(y^{(n-1)})$,

$$\|\dot{\phi}_{\delta_k} Nf - Sf\| < \epsilon + \gamma. \tag{6.10}$$

We now begin to define the operator $N$ for all $f$ in

$$T_k(y^{(n-2)}) \equiv N_{\delta_k}\{y = (y^{(n-2)}, y_{n-2}, y_n): y_{n-1}, y_n \in \mathbb{R}\}.$$

First note that now $N$ is defined for all $f \in T_k(y^{(n-1)})$, as long as $y^{(n-1)} \in \mathcal{Q}_n$. In order to extend this definition, we note that for $k$ sufficiently large, (8.19)(6.10) holds for all $f \in T(y^{(n-1)})$ for $y^{(n-1)}$ fixed. Thus given $y^{(n-2)} \in \mathcal{Q}_{n-2}$, then for an arbitrary $\rho > 0$ there exists a $k$ such that the set of $y_{n-1} \in \mathbb{Q}_{n-1}(y^{(n-2)})$ with (8.18)(6.9) holding (with its third argument replaced by $(y^{(n-2)}, y_{n-1}, y_n)$ and $\gamma$ replaced by $2\gamma$) for all $y_n$ is $\rho$-dense in the set

$$P_{n-1} \equiv \{z: (y^{(n-2)}, z) \in NT(y^{(n-1)})\}.$$

By $\rho$-dense here we mean that for every $z \in P_{n-1}$, there exists a $z_1 \in P_{n-1}$ satisfying

$$e(N, \dot\phi_{\delta_k}, (y^{(n-2)}, z_1, y_n)) < \epsilon + 2\gamma \quad \text{for all } y_n \in \mathbb{R}$$

such that $|z - z_1| < \rho$. Again using continuity arguments (namely the uniform continuity of the algorithms $\phi_{\delta_k}$), we conclude that for sufficiently large $k$ (and thus sufficiently small $\rho$) above, we can replace the information operator $N(y^{(n-2)}), \cdot)$ by an operator $N_1(y^{(n-2)}, \cdot)$ such that

$$e(N_1, \dot\phi_{\delta_k}, (y^{(n-2)}, y_{n-1}, y_n) < \epsilon + 2\gamma$$

for all $(y_{n-1}, y_n)$. To see this, note that for any $y^{(n-2)} \in \mathcal{Q}_{n-2}$, and $y_{n-1}$ such that (8.22) holds for all $y_n$, we have

$$e(N, \dot\phi_{\delta_k}, (y^{(n-2)}, y_{n-1}, y_n)) \equiv \sup_{f \in T(y^{(n-2)}, y_{n-1}, y_n} \|\dot\phi_{\delta_k} Nf - Sf\| < \epsilon + 2\gamma.$$

Again using the uniform continuity of the $\dot\phi_{\delta_k}$, there is a $\rho_1 > 0$ such that for $|y'_{n-1} - y_{n-1}| \le \rho_1$, we have for each $y_n$

$$\sup_{f \in T(y^{(n-2)}, y'_{n-1}, y_n)} \|\dot\phi_{\delta_k}(y^{n-2}, y_{n-1}, y_n) - Sf\| < \epsilon + 2\gamma; \tag{6.11}$$

note that in the above supremum we have taken $f$ from a set $T$ whose argument uses a different choice of $y_{n-1}$ than the algorithm $\dot\phi_{\delta_k}$ in the supremum. This follows again because we can restrict this supremum to a finite dimensional transverse subspace $V$, such that the maximal set distance of $S(V \cap F_0)$ from $SF_0$ is arbitrarily small. Since (6.11) above is true for all $\rho_1 > 0$, we can select $k$ so that for every $y_{n-1} \in \mathbb{Q}(y^{(n-2)})$ there is a $y'_{n-1}$ such that $|y_{n-1} - y'_{n-1}| < \rho_1$, and $e(N, \dot\phi_{\delta_k}, (y^{(n-2)}, y'_{n-1}, y_n)) < \epsilon + 2\gamma$ for all $y_n \in \mathbb{R}$. Therefore, together with (6.9) this shows that we can for a given choice of $\rho_1$ redefine the operator $N$ on $T(y^{(n-2)})$ to obtain a new operator $N_1$ such that:

*(i)* For $f \in T(y^{(n-2)}, y_{n-1})$ such that $e(N, \dot\phi_{\delta_k}, (y^{(n-2)}, y_{n-1}, y_n)) < \epsilon + 2\gamma$ for all $y_n \in \mathbb{R}$, we choose $N_1 f = N f$.

*(ii)* For $f \in T(y^{(n-2)}, y_{n-1})$ otherwise, choose (in a measurable way) a $y'_{n-1}$ such that the condition in *(i)* holds with $y_{n-1}$ replaced by $y'_{n-1}$, and define
$$N_1(y^{(n-2)}, y_{n-1}, y_n) = N(y^{(n-2)}, y'_{n-1}, y_n).$$

In this way, the operator $N_1$ has the property that
$$e(N_1 \dot\phi_{\delta_k}, (y^{(n-2)}, y_{n-1}, y_n)) < \epsilon + 2\gamma$$
for all choices of $(y_{n-1}, y_n)$.

We continue araguing in this way. For $y^{(n-3)} \in \mathcal{Q}_{n-3}$, it follows again that for $k$ sufficiently large, the set of $y_{n-2}$ such that
$$e(N_1, \dot\phi_{\delta_k}, (y^{(n-3)}, y_{n-2}, y_{n-1}, y_n)) < \epsilon + 3\gamma \tag{6.12}$$

is eventually $\rho$-dense for any choice of $\rho$, so that by the same continuuity argument as before, we can replace $N_1$ by a redefined operator $N_2$ which is defined for all $y = (y^{(n-3)}, y_{n-2}, y_{n-1}, y_n)$, and which satisfies (6.12) for all $(y_{n-2}, y_{n-1}, y_n) \in \mathbb{R}^3$. We continue redefining $N$ in this way, at each stage, getting for each fixed $y^{(n-l)} \in \mathcal{Q}_{n-l}$ an operator $N_{l-1}$ defined for all $y = (y^{(n-l)}, y_{n-l+1}, y_{n-l+2}, \ldots, y_n)$ for arbitrary $(y_{n-l+1}, y_{n-l+2}, \ldots, y_n)$ with the property that
$$e(N_{l-1} \dot\phi_{\delta_k}, (y^{(n-l)}, y_{n-l+1}, y_{n-l+2}, \ldots, y_n)) < \epsilon + l\gamma;$$

we may need to increase the range of $k$ for which our reasoning is valid at each stage of this process; however, since at each stage we assume that our identities hold for all $k$ larger than some fixed $K$, increasing the lower threshold $K$ will not change the validity of previous assertions.

Finally, at the last stage we have an operator $N_{n-2}(y)$ obtained as above, and we replace it by an operator $\tilde{N}_\gamma(y) = N_{n-2}(c(\gamma)y)$ such that $\tilde{N}_\gamma$ is defined for all $y$, and such that for sufficiently large $k(\gamma)$,
$$e(\tilde{N}_\gamma, \dot\phi_{\delta_{k(\gamma)}}, y) < \epsilon + l\gamma + M(\ell\gamma)$$

for all $y$. Above, $c(\gamma)$ is a multipllicative constant satisfying $c(\gamma) \xrightarrow[\gamma \to 0]{\uparrow} 1$, and chosen so that $\tilde{N}_\gamma$ is defined for all arguments $f$, and $M(\gamma)$ is the modulus of continuity of the family $\phi_{\delta_k}$, which has been chosen to be uniformly continuous, with $\ell$ a constant. The term $M(\ell\gamma)$ is needed to adjust for the fact that $N_2$ has been replaced by $\tilde{N}_\gamma = N_2(c(\gamma) \cdot )$ (note that by our assumption $M(\gamma) \xrightarrow[\gamma \to \infty]{} 0$).

It follows that the $\epsilon + l\gamma$- worst case complexity is bounded above by

$$C(\tilde{N}_\gamma, \dot{\phi}_{\delta_{k(\gamma)}}) = cn + \text{cost}(\tilde{N}_\gamma, \dot{\phi}_{\delta_{k(\gamma)}}) \leq \text{comp}_{\delta_k}(\epsilon) + \delta_k$$

using (8.4). Letting $k \to \infty$, we conclude that

$$\text{comp}^{\text{wor}}(\epsilon) \leq \lim_{k \to \infty} \text{comp}_{\delta_k}(\epsilon) + \delta_k = \lim_{k \to \infty} \text{comp}_{\delta_k}(\epsilon) = \lim_{\delta \to 0} \text{comp}_\delta(\epsilon),$$

so that

$$\text{comp}^{\text{wor}}(\epsilon) = \lim_{\delta \to 0} \text{comp}_\delta(\epsilon),$$

as desired. This proves asertion *(i)*.

The second assertion is proved in the remarks below. □

**Remarks:**

**1.** To prove assertion *(ii)*, we first consider the removal of the assumption of bounded complexity. If the complexity $\text{comp}^{\text{wor}}(\epsilon)$ fails to be bounded for small $\epsilon$, the following example shows that the conclusion of Theorem 12 is false.

For a given information operator $N$ and algorithm $\phi$, again define the worst case error by

$$e^{\text{wor}}(N, \phi) = \sup_{f \in F_0} \|\phi(N(f)) - Sf\|,$$

and the $\delta$-probabilistic error (for a given measure $\mu$ on $F_0 \subseteq F$) by

$$e_\delta(N, \phi) = \inf_{\mu(A_\delta) \geq 1-\delta} \sup_{f \in A_\delta} \|\phi(N(f)) - Sf\|,$$

where it is understood here and afterward that all sets $A_\delta$ in such infima are contained in $F_0$.

Recall we are assuming here a model of computation in which the complexity is defined by

$$C_\delta(\epsilon) = \inf_{\phi \in \Phi} \{c \text{ card } N + \sup_{f \in F_0} \text{cost}(\phi, Nf) : e_\delta(N, \phi) \leq \epsilon\},$$

where card $N$ denotes the cardinality of the information operator $N$, and cost $(\phi, Nf)$ is the assumed cost of the algorithm $\phi$ applied to information $Nf$. A similar definition (using $e^{\text{wor}}$ instead of $e_\delta$) is valid for the complexity $C^{\text{wor}}(\epsilon)$. Under some mild assumptions regarding the model of cost (i.e., on how $\text{cost}(\phi, Nf))$ depends on $\phi$), it has been shown above that as $\delta$ converges to 0,

$$C_\delta(\epsilon) \xrightarrow[\delta \to 0]{} C^{\text{wor}}(\epsilon). \tag{6.13}$$

However, there exist models in which $\text{cost}(\phi, y)$ depends continuously on $\phi$ and $y$, for which (9.1)(6.13) fails. In fact, such models exist in canonical situations, involving Hilbert space with Gaussian measure.

We will construct such an example. Let $F$ be a Hilbert spacae with Gaussian measure, $F_0$, be its unit ball of $F$, $S : F \to F$ denote the identity operator. Let $\mu$ be a Gaussian measure on $H$. Note that here the Gelfand $n$-widths of $SF_0$ do not go to 0. Let $N$ be an adaptive information operator of cardinality $n$, and $\phi$ be a corresponding algorithm, for which $e(N, \phi) \leq \epsilon$, and

$$C(N, \phi) \equiv c \operatorname{card} N + \sup_{f \in F_0} \operatorname{cost}(\phi, Nf)$$

is close to its minimal value $C^{\text{wor}}(\epsilon)$. We will here assume a model of computation for which it is true that for all $y$, $\operatorname{cost}(\phi, y)$ is very high unless $\phi(y)$ is near the boundary of $N^{-1}(y) \cap F_0$ (this case can easily be made precise). Then it is clear that (independent of the cardinality of $N$) the worst case error (given a maximum cost we are willing to bear) of our information operation together with the algorithm can be made arbitrarily close to 2. This is because the optimal estimate $\phi(0)$ of an $f \in F_0$ in the kernel of $N$ will be close to the boundary of $F_0$ (by the assumption above on the model of cost), so that $f$ can be chosen in a way that $\|\phi(0) - f\|$ is close to 2. Thus we can arrange things so that

$$e^{\text{wor}}(N, \phi) \geq 2 - \gamma,$$

for any $\gamma > 0$.

On the other hand, we can show that the probabilistic error $e_\delta(N, \phi_\delta)$ of certain other algorithms in our model is small, and in fact $1/\sqrt{2}$ of the error of $e^{\text{wor}}(N, \phi)$, even as $\delta \to 0$. To see this, note that due to the fact that the measure $\mu$ and thus its conditional measure $\mu_y$ on $N^{-1}(y)$ is Gaussian, there exist sets $A_\delta$ of measure arbitrarily close to 1 such that for each $y$ it is true that for some $z_y \in N^{-1}(y)$, $a \in A_\delta \cap N^{-1}(y)$ implies $|\langle z_y, a \rangle| \leq \gamma$. Intuitively, the sets $A_\delta$ have slices in $N^{-1}(y)$ which are very thin in some directions (given by $z_y$) in $N^{-1}(y)$. Thus using the same information operator $N$ as above, choose $\phi_\delta(y) = z_y$. Then it is easy to check that (if $\epsilon$ is chosen to be small)

$$e_\delta(N, \phi_\delta) = \sqrt{2} + \eta,$$

where $\eta \xrightarrow[\gamma \to 0]{} 0$. This follows from the fact that

$$e_\delta(N, \phi_\delta) = \sup_{a \in A_\delta \cap N^{-1}(y)} \|a - z_y\|,$$

together with the fact that $|\langle z_y, a \rangle| \leq \eta$ (where $\eta$ is small), and that both $z_y$ and $a$ are in the unit ball.

Our conclusion is that for a given cost $k$ and $\alpha > 0$ ($\alpha$ small), we can find information operator $N$ and algorithm $\phi$ such that

$$C(N, \phi) = C,$$

while

$$e^{\text{wor}}(N, \phi) \leq e^{\text{wor}}(C) + \alpha = 2 + \alpha$$

where

$$e^{\text{wor}}(C) \equiv \inf_{N, \phi} \{e^{\text{wor}}(N, \phi) : C(N, \phi) \leq C\}.$$

However, with this same information operator and the proper choice of $\phi_\delta$ for each $\delta$, we find that $C(N, \phi_\delta) = C$, while

$$e_\delta(\mathrm{N},\phi_\delta) = \sqrt{2} + \eta,$$

with $\eta \xrightarrow[\gamma \to 0]{} 0$.

We conclude that in this model, for $\epsilon = \sqrt{2} + \beta$ for some positive $\beta$, $C^{\mathrm{wor}}(\epsilon)$ can be kept arbitrarily high (since by our assumption a central algorithm $\phi$ is costly), while $C_\delta(\epsilon) = c$ card $N +$ cost$(N, \phi_\delta)$ can in fact be kept low for all $\delta$, since we may here use a $\phi_\delta$ which gives points $\phi_\delta(y)$ in the boundary of the set $N^{-1}(y) \cap F_0$, which by assumption are easy to compute, and still keep our error less than $\sqrt{2} + \beta$.

**2.** Second, we will show that the removal of the assumption of existence of $\phi_\delta$ which are uniformly continuous makes statement *(i)* of the theorem false. We give a sketch here. More realistic examples exist, but this is a simple one which illustrates the principle on which our model of computation is based. Consider the situation in which $F$ is two dimensional, $F_0$ is the Euclidean unit ball in $F$, and $S$ is the identity operator on $F$. Assume that $\mu$ is Lebesgue measure on $F_0$, normalized to have unit measure. Further, assume a model of cost in which $\phi$ is discontinuous, in that for any information operator $N$, the complexity of computing a good approximation $\phi(Nf)$ with $Nf = 1/2$ is prohibitively high (say a cost of $M >> 0$; recall in our model of computation and in the above theorem we assume that the cost $\phi(y)$ may depend in any way on $y$ and on $N$; dependence on $N$ is a natural assumption in that different choices of $N$ yield different domains for the function $\phi(y)$. Assume further that for any other value of $Nf$, the cost of $\phi$ is some small number $m << 1$. In this case, given $\epsilon = 1/2$ and any information operator $N$, the worst case $\epsilon$-complexity is $c + M$ if the cardinality of the best information operator $N$ is 1 (i.e., the cost $c$ of one computation plus the cost of computing $\phi$ if $Nf = 1/2$). If we use the best information operator of cardinality 2, then the worst case cost is $2c + 0$, since in that case $\phi$ is the identity operator and so has 0 cost. So in any case, the worse case cost is

$$\mathrm{comp}^{\mathrm{wor}}(\epsilon) = \min(c + M, 2c).$$

However, the $\delta$-probabilistic complexity is smaller, since the set of $f$ for which $Nf = 1/2$ has measure 0. In this case for all $\delta \geq 0$, the complexity is

$$\mathrm{comp}_\delta(\epsilon) = \min(c + m, 2c).$$

Therefore, if we assume that $m < c < M$, we have that

$$\lim_{\delta \to 0} \mathrm{comp}_\delta(\epsilon) = c + m < \mathrm{comp}^{\mathrm{wor}}(\epsilon) = c + M,$$

showing the conclusion of Theorem 8.1 to be false in this case.

# 7. Semicontinuity of the Adaptive Gelfand Radius

The techniques of proof of Lemma 3.4 and Theorem 3 lead to another interesting result, namely Theorem 12, regarding the lower semicontinuity of the so-called adaptive Gelfand radius worst case complexity. In addition to continuity results on the radius of information (Proposition 5, Corollary 2.2 ), it can be shown that for a fixed $n$, the adaptive Gelfand $n$-radius of a set $N^{-1}(y)$, where $N$ is a nonadaptive operator of fixed cardinality, is a lower semicontinuous function of $N$. That is, if we "slice" a convex set $A$ with the hyperplane $N^{-1}(y)$, then not only is the radius of this set continuous in $N$ and $y$, but the adaptive Gelfand radius of this slice is lower semicontinuous in the same variables. The technique of this proof involves the finite dimensional reduction used in the proof of Theorem 3.

*Proof of Theorem 12:* Assume this is false. Then there exists a pair $(N, y)$ with $(N_k, y_k) \xrightarrow[k \to \infty]{} (N, y)$ in our topology, such that $R(A_{k y_k}) \xrightarrow[k \to \infty]{} R(A_y)$, where

$$A_{k y_k} = N_k^{-1}(y_k) \cap A; \quad A_k = N^{-1}(y) \cap A.$$

Let $M_k$ be adaptive information operators of order $n$ such that

(9.4) $\qquad \operatorname{Rad}(A_{k y_k}) < R(M_k, A_{k y_k}) < \operatorname{Rad}(A_{k y_k}) + \epsilon_k,$ $\qquad$ (7.1)

with

$\epsilon_k \xrightarrow[k \to \infty]{} 0$. As in previous proofs (see proof of Lemma 3.4), we can without loss of generality take a subsequence such that for a countable set $\mathcal{Q}$ of information elements $z$, we have $K_{ki}(z, \cdot) \xrightarrow[k \to \infty]{} K_i(z, \cdot)$, where $K_{ki}$ denotes the $i^{th}$ component linear functional of $M_k$, with convergence in the weak topology. Let $M(z, \cdot) = (K_1(\cdot), K_2(x, \cdot), \ldots, K_n(z, \cdot))$. Since $\mathcal{Q}$ is dense in $\mathbb{R}^n$, we can choose a $z \in \mathcal{Q}$ such that for given $\eta > 0$,

$$R(M, z, A_y) > R(M, A_y) - \eta. \qquad (7.2)$$

Now we claim that

$$R(M_k, z, A_{k y_k}) \xrightarrow[k \to \infty]{} R(M, z, A_y).$$

To see this, consider the combined (nonadaptive) operators $P_k = (N_k, M_k(z, \cdot)) = (L_1, \ldots, L_d; K_1, K_2(z, \cdot), \ldots, K_n(z, \cdot))$. We have $P_k \xrightarrow[k \to \infty]{} P = (N, M(z, \cdot))$ in the weak operator topology. We also have by Proposition 5

$$R(M_k, z, A_{k y_k}) = R(P_k, (y_k, z), A) \xrightarrow[k \to \infty]{} R(P, (y, z), A) = R(M, z, A_y), \qquad (7.3)$$

as desired. Note that the convergence in (7.3) is in the weak operator topology, however, Proposition 5 applies because by Lemma C we can restrict $A$ and $P$ to a finite dimensional affine subspace $T$, for which (using the notation of the proof of Lemma 4.4)

$$R(P, (y, z), A_T) > R(P, (y, z), A) - \epsilon \tag{7.4}$$

for arbitrarily small $\epsilon > 0$. If the rest of the radii evaluations in (7.3) are similarly restricted to $T$, then follows strictly from Proposition 5 (since in $T$ the convergence of $P_k$ to $P$ is in the uniform operator topology). However, it is easy to check that if a sequence of $T$ are chosen so that $\epsilon \to 0$ in (7.4), then (7.3) is verified.

By (7.1) and (7.2) therefore,

$$\begin{aligned}
\liminf_{k \to \infty} R(A_{k y_k}) &= \liminf_{k \to \infty} R(M_k, A_{k y_k}) \\
&\geq \liminf_{k \to \infty} R(M_k, z, A_{k y_k}) \\
&= R(M, z, A_y) \\
&\geq R(M, A_y) - \eta \\
&\geq R(A_y) - \eta
\end{aligned}$$

Letting $\eta \to 0$, we conclude that

$$\liminf_{k \to \infty} R(A_{k y_k}) \geq R(A_y),$$

or

$$\liminf_{k \to \infty} R_n(N_k . y_k) \geq R_n(N, y). \tag{7.5}$$

Since (7.5) holds for a subsequence of our original sequence, it also holds for the original sequence, and thus in general for any sequence.

Thus we conclude that for any $\alpha$, if $R_n(N_k, y_k) \leq \alpha$, then $R_n(N, y) \leq \alpha$ so that $\{(N, y) \,|\, R_n(N, y) \leq \alpha\}$ is closed, showing that $R_n(N, y)$ is lower semicontinuous. $\square$

**Remark:** Though it is likely that in fact the above Gelfand radius is a continuous function of $N$ and $y$, the above method of proof does not prove upper semicontinuity. It is likely that a modification of the arguments in Proposition 4 would prove this, and thus full continuity of the adaptive Gelfand radius.

**Acknowledgments:** The author would like to thank S. Heinrich and H. Wozniakowski for extensive and helpful discussions on this subject.